\documentclass[12pt,twoside]{article}

\usepackage{hyperref}
\usepackage{setspace}
\usepackage{indentfirst}
\usepackage{geometry}
\usepackage[table,xcdraw]{xcolor}
\usepackage{fancyhdr}
\usepackage{caption}
\usepackage[labelfont=bf]{caption}
\usepackage{natbib}
\usepackage{graphicx}
\usepackage{floatrow}
\usepackage{tabularx}
\usepackage{multirow}
\usepackage{titling}
\usepackage{enumitem}
\usepackage{booktabs}
\usepackage[hang,flushmargin]{footmisc}
\usepackage{titlesec}
\titleformat{\section}{\normalfont\bfseries\filcenter}{}{0pt}{}
\titleformat{\subsection}{\normalfont\bfseries\filcenter}{}{0pt}{\itshape}
\titleformat{\subsubsection}{\normalfont\bfseries\filcenter}{}{0pt}{\itshape}
\titlespacing*{\section}
  {0pt}{-.1\baselineskip}{-.1\baselineskip}  
\titlespacing*{\subsection}
   {0pt}{-.1\baselineskip}{-.1\baselineskip}  
\usepackage[T1]{fontenc}
\usepackage{array} 
\usepackage{siunitx}
\usepackage{dcolumn}
\usepackage{subcaption}
\usepackage{ragged2e}

\newcolumntype{L}{D{.}{.}{-3}}

\date{}
\providecommand{\keywords}[1]
{
   \small	
  \textit{\hspace{-1em} Keywords: } #1
}

\fancypagestyle{firstpage}{
  \fancyhf{}
  \fancyfoot[L]{ \footnotesize Copyright \copyright 2024 (Mendelsohn, Vijan, Card, and Budak). Creative Commons Attribution-NonCommercial-NoDerivatives 4.0 International Public License. Available at: \url{http://journalqd.org}} 
  \lhead{ \footnotesize Journal of Quantitative Description: Digital Media 4(2024), 1–61} 
\rhead{ \footnotesize 10.51685/jqd.2024.icwsm.9}
}
\title{\normalsize \textbf{Framing Social Movements on Social Media:\\
Unpacking Diagnostic, Prognostic, and Motivational Strategies} \vspace{-1.5em}} 

\author {\begin{tabular}{c}
\normalsize JULIA MENDELSOHN\\
\normalsize MAYA VIJAN \\
\normalsize DALLAS CARD \\
\normalsize CEREN BUDAK \\ 
\normalsize University of Michigan, USA
\end{tabular}
\thanks{\hspace{-1em} Julia Mendelsohn: juliame@umich.edu}
\thanks{\hspace{-1em} Maya Vijan: vijanm@umich.edu}
\thanks{\hspace{-1em} Dallas Card: dalc@umich.edu}
\thanks{\hspace{-1em} Ceren Budak: cbudak@umich.edu}
\thanks{\hspace{-1em} Date submitted: 2023-09-29}}

\usepackage{lipsum}
\renewcommand\footnotemark{}
\begin{document}
\maketitle
\thispagestyle{firstpage}
\vspace{-9em}
\begin{abstract}
   Social media enables activists to directly communicate with the public and provides a space for movement leaders, participants, bystanders, and opponents to collectively construct and contest narratives. Focusing on Twitter messages from social movements surrounding three issues in 2018-2019 (guns, immigration, and LGBTQ rights), we create a codebook, annotated dataset, and computational models to detect diagnostic (problem identification and attribution), prognostic (proposed solutions and tactics), and motivational (calls to action) framing strategies. We conduct an in-depth unsupervised linguistic analysis of each framing strategy, and uncover cross-movement similarities in associations between framing and linguistic features such as pronouns and deontic modal verbs. Finally, we compare framing strategies across issues and other social, cultural, and interactional contexts. For example, we show that diagnostic framing is more common in replies than original broadcast posts, and that social movement organizations focus much more on prognostic and motivational framing than journalists and ordinary citizens.  
\end{abstract}

\vspace{-1em}
\keywords{\textit{framing, social movements, social media, Twitter} \vspace{8ex}}
\vspace{-1em}

Social movements use social media to draw attention to their cause, disseminate information, coordinate offline collective action, and build collective identity \citep{harlow_social_2012,jost_how_2018}. Given its prominent role, social media ought to be studied in the context of contemporary social movements, even if we cannot make outright causal claims about the effects of social media use on movement success \citep{kidd_social_2016}. At the core of social movements are networks of interactions between people with shared identities and goals \citep{diani_concept_1992}; it is through these interactions that people collectively make sense of the world and the role of their movements in that world. We focus on this aspect of social movement communication, collective action framing, to characterize the discursive construction of social movements across three issue areas on Twitter from 2018-2019.\footnote{Twitter rebranded as X in July 2023. However, we use the terms ``Twitter'' and ``tweets'' throughout this paper as they accurately describe the platform during the time period that we study.}  

\textit{Framing} has diverse conceptualizations and operationalizations across disciplines, particularly cognitive linguistics, psychology, communication, and sociology \citep{cacciatore2016end}. Primarily situated within sociology, social movement scholarship has largely adopted \cite{goffman1974frame}'s definition of frames as ``schemata of interpretation'' that help people ``locate, perceive, identify, and  label'' new information about the world around them \citep{snow_frame_1986,van_dijk_analyzing_2023}. Collective action frames are further strategic aspects of communication, ``intended to mobilize potential adherents and constituents, to garner bystander support, and to demobilize antagonists'' \citep{snow_ideology_1988}. We draw upon Benford and Snow's typology of \textit{core framing tasks} \citep{snow_ideology_1988, benford_framing_2000}. \textit{Diagnostic} framing involves identifying problems, their causes, and who to blame or hold responsible. \textit{Prognostic} framing involves articulating proposed solutions, plans of attack, and strategies or tactics for carrying out those plans. \textit{Motivational} framing refers to motivating people to participate in the social movement through calls to action. This paper addresses the broad question: how do people accomplish diagnostic, prognostic, and motivational core framing tasks on Twitter?

We analyze a dataset of 1.85M tweets from movements in 2018-2019 surrounding three issue areas: \textit{guns}, \textit{immigration}, and \textit{LGBTQ rights}, primarily within the U.S. context. We first develop a codebook, which we use to manually annotate a sample of 6,000 tweets for relevance, stance, and diagnostic, prognostic, and motivational framing strategies. We then use supervised machine learning techniques to infer labels for the rest of the tweets in our dataset. By enabling us to analyze framing at scale, our automated data labeling approach facilitates a more comprehensive description of social movement discourse on Twitter than small-N samples. We then conduct a fine-grained linguistic analysis of each core framing task to better understand how social movements create meaning from raw textual material \citep{vicari_measuring_2010}. We show the prevalence of moral language across issue areas and core framing tasks, and highlight how core framing tasks differ in their personal pronoun usage, suggestive of boundary framing processes \citep{snow_frame_1986}.

The resulting large-scale labeled dataset allows for the investigation of fine-grained temporal patterns and comparisons of framing across many cross-sections of data. In particular, we explore how attention to each core framing task varies across a set of movement-level sociocultural dimensions, including issue area, ideology, and offline protest activity levels. We also compare framing across several message-level interactional factors, namely the author's role (as a journalist, social movement organization, or member of the public) and tweet type (traditional ``broadcast'' tweets versus quote tweets and replies). Our results suggest a degree of stability in framing practices, especially over time and within the same ideology across issues. At the same time, we uncover significant frame variation across several factors. For example, social movement organizations employ far more \textit{prognostic} and \textit{motivational} framing than their journalist and general public counterparts. We also show substantial variation across tweet types, with replies and quote tweets being much more likely to contain \textit{diagnostic} framing than broadcasts. Taken altogether, our work emphasizes both the importance of message context and the value of a consistent methodology when studying social movement communication.\footnote{Our dataset, annotation codebook, code, and models are all available at: \url{https://github.com/juliamendelsohn/social-movement-framing/}}

\section{Background}

We first introduce the framing perspective in social movement studies and discuss the interplay between social media affordances and social movement communication. We specifically highlight prior work that analyzes movements' engagement with diagnostic, prognostic, and motivational core framing tasks. We then motivate our selection and investigation of frame variation and alignment across multiple sociocultural dimensions, including stance, offline events, author roles, and audiences. Next, we provide a brief overview of the three issue areas that we study: \textit{gun policy/violence}, \textit{immigration}, and \textit{LGBTQ rights}. We conclude this section with a review of computational approaches to frame analysis. 

\textbf{Framing and Social Movements.} Social movements are sustained efforts to enact or hinder social and political changes \citep{jasper_protest_2014}, and are characterized by networks of interactions between individuals and groups united by shared collective identities \citep{diani_concept_1992, della_porta_social_2006}. Since the 1980s, scholars have embraced framing as a cultural perspective to complement predominant theories of resource mobilization and political opportunity structure \citep{snow_frame_1986}. Framing refers to the dynamic, interactive process of constructing and negotiating shared meaning of a movement \citep{snow_frame_1986}, ``involving various actors within a discursive context consisting of movement activists, actual and potential adherents, countermovement proponents, social control agents, government and social policy agents, the media, and often a watchful public'' \citep{snow_imperious_2023}. Effective framing is associated with protest participation and movement success \citep{cress_outcomes_2000,della_porta_social_2006,somma_what_2019}. 

Social movement research has traditionally taken interpretive approaches to analyze framing practices on a macro-scale, at the group or movement-level \citep{gerhards_mesomobilization_1992}. However, critics both within and outside the field have called for more empirical studies \citep{benford_insiders_1997,snow_emergence_2014}, and advocate for a greater focus on micro-level frame analysis within interactions on both theoretical and methodological grounds \citep{johnston_methodology_1995,johnston_methodology_2013,hedley_microlevel_2007,vicari_measuring_2010,vicari_frame_2023,gordon_framing_2023,van_dijk_analyzing_2023}. Methodologically, micro-frame analysis of individual texts ``enables the researcher to speak about frames with a great deal more empirical grounding'' \citep{johnston_methodology_1995}. Theoretically, movement-level frames emerge through the aggregation, contestation, layering, and transformation of framing within individual messages or interactions \citep{hedley_microlevel_2007}. Analysis at the micro-level sheds light on the dynamic processes associated with framing, and helps ``illuminate moment-by-moment constructions of meanings, identities, and relationships'' \citep{gordon_framing_2023}.

\textbf{Social Movements and Social Media.}
Social media has upended many traditional assumptions about collective action \citep{kavada_social_2016}. In contrast to the high organizational structure, emphasis on collective identity, and rigid collective action frames in offline collective action, \citet{bennett_digital_2011,bennett_logic_2012} argue that social media engenders a new logic of ``connective action'' based on personalization of content and frames to mobilize different audiences. It facilitates a hybrid environment in which activists, bystanders, news media, and politicians all participate and interact to negotiate meanings of collective actions \citep{pavan_embedding_2014,barnard_tweeting_2018}. \cite{meraz_networked_2013} articulate ``networked framing'' in such environments as a process by which frames are ``persistently revised, rearticulated, and redispersed by both crowd and elite'' and ought to be interpreted in the context of the ambient conversations taking place concurrently on the platform.

Marked by this hybridity and decentralized communication, social movement scholarship has interrogated connections between social media activism and shifts in the distribution of power \citep{dimond_hollaback_2013}. Movements may be able to gain public attention (a key resource for mobilization) and share their own narratives without relying on traditional gatekeepers such as mass media \citep{tufekci_not_2013,mundt_scaling_2018,molder_framing_2022}. There may also be shifts in power within social movements themselves. For example, scholars debate the role of centralized formal social movement organizations (SMOs) in online protest \citep{earl_future_2015,spiro2016shifting,bozarth_beyond_2021}. Some argue that leadership still exists in online social movements, but takes a different shape, where opinion leaders may include not only SMOs, but also ordinary citizens, celebrities, and community administrators \citep{poell_protest_2016,gerbaudo_social_2017,pond_riots_2019,liang_opinion_2021,uysal_social_2022}.

Social movements have leveraged platform affordances in many other ways, such as for building collective identity \citep{harlow_social_2012,milan_when_2015,flores-saviaga_mobilizing_2018}. Platforms such as Twitter facilitate forming connections between people with shared interests and values who may have no geographic proximity or other offline ties with each other \citep{tremayne_anatomy_2014,van_haperen_swarm_2023}. Through both algorithmic curation and crowd collaboration, social media affordances enable movement actors to consume, share, recommend, and filter information \citep{starbird_how_2012,meraz_networked_2013,tillery_what_2019,etter_activists_2021}, including coordination of offline protest \citep{harlow_social_2012,earl_this_2013,jost_how_2018,tsatsou_social_2018}. Prior work has also analyzed framing through the use of technological features such as hashtags \citep{meraz_networked_2013,ince_social_2017} and reposts \citep{nip_networked_2016}.

A much smaller body of work has coded and described diagnostic, prognostic, and motivational framing strategies in social media posts to analyze a variety of movements, primarily focusing on prominent activists and organizations. Using \citet{benford_framing_2000}'s definition of these core framing tasks, \citet{vu_social_2021} analyze framing by climate change organizations on Facebook and \citet{molder_framing_2022} qualitatively analyze youth climate activist Greta Thunberg's Instagram posts. \citet{hon_social_2016} conducts a qualitative analysis of the Facebook page from the Million Hoodies racial justice movement to identify how the movement engages with each core framing task. \citet{phadke_framing_2018} and \citet{phadke_many_2020} compare how hate groups frame their hate-based movements on Facebook and Twitter by developing a fine-grained typology of domain-specific frame components within diagnostic, prognostic, and motivational framing. In the context of protest against the Singaporean government's immigration policy changes, \citet{goh_protesting_2016} compare organizers' and protestors' framing strategies in blogs and Facebook posts. The authors develop sub-categories within each framing task through exploratory analysis guided by \citeauthor{benford_framing_2000}'s definitions, such as coding sub-categories of the diagnostic frame for problem, victim, and causal agent identification. The current study differs substantially from these works in domain, methodology, scale, and broad research questions. Nevertheless, they inspire us to pursue a similar strategy in creating our codebook, particularly in identifying several sub-categories for diagnostic and prognostic framing.






 \textbf{Frame Variation.} Despite many studies on social movement framing, little attention has been dedicated to empirical analyses of frame variation across sociocultural contexts \citep{snow_emergence_2014}. Given the importance of moving beyond single case studies and pursuing comparative work in social movement studies \citep{tarrow1996social}, we heed \cite{snow_emergence_2014}'s call for comparisons of framing across movements, actors, and time, and here briefly motivate the sociocultural dimensions that we study. 

Movements co-exist and engage with other movements, and the same actors may participate in multiple movements across different issue areas \citep{carroll_master_1996}. Cultural practices, including framing, have also been shown to diffuse both within and across movements \citep{soule_diffusion_2018}. Movements also exist in direct competition with opposing countermovements. As one of the primary goals of collective action framing is to ``demobilize antagonists'' \citep{snow_ideology_1988}, frames develop over the course of sustained interaction with countermovements \citep{ayoub_movementcountermovement_2020}. Movement and countermovement frames may be closely-aligned, especially when drawing from the same cultural themes \citep{ayoub_movementcountermovement_2020,sun_national_2023}. At the same time, movements may develop frames to challenge or contest the opposition's \citep{mccaffrey_competitive_2000,stewart_drawing_2017}. Framing may also vary due to differences in movement and countermovement opportunities and constraints; for example, \textit{gun rights} organizations' financial advantages over \textit{gun control} organizations allow them to rely less on ``attention-grabbing moments'' \citep{laschever_growth_2021}. Framing processes are closely intertwined with collective action activity. Not only are frames deployed in order to mobilize collective action participation \citep{benford_framing_2000}, but they also evolve during cycles of protest \citep{snow_master_1992}. Framing practices further shift in response to protest waves or even individual collective action events \citep{ellingson_understanding_1995, swart_league_1995, valocchi_riding_2006}.

Social movement discourses on online platforms, particularly hybrid media environments such as Twitter, are characterized by the presence of and interactions between many groups of stakeholders, including social movement organizations (SMOs), journalists, and ordinary citizens \citep{meraz_networked_2013, jackson_ferguson_2016, caren_contemporary_2020, isa_social_2018, hunt_horizontal_2023}. These stakeholder groups each have different goals and complex relationships with each other, through which framing takes center stage. On one hand, SMOs seek to gain media attention and influence news coverage \citep{andrews_making_2010,gibson_effects_2023}. On the other hand, news media frequently offers delegitimizing frames of protestors to uphold the status quo through the ``protest paradigm'' \citep{mcleod_framing_1999}. However, more recent work has found news media to also offer more sympathetic and legitimizing coverage that considers protestors' grievances and demands \citep{mourao_framing_2021,gruber_troublemakers_2023}. In digitally-mediated social movements, journalists have even found themselves alongside SMOs as core movement actors who mediate the flow of relevant information \citep{isa_social_2018}.

The role of SMOs in online movements remains unclear. Given the centrality of SMOs in pre-digital movements, a major focus of earlier framing research has been in frame alignment between SMOs and the populations they seek to mobilize \citep{snow_frame_1986}. As social media offers more opportunities for individual citizens to become opinion leaders \citep{meraz_networked_2013, jackson_ferguson_2016}, the utility and necessity of SMOs has come into question \citep{earl_new_2002,earl_future_2015}. At the same time, SMOs remain a central source of information, offer credibility and legitimacy, have greater capacity to organize offline collective action, and may even be more successful in collective identity formation and recruitment online than their non-organizational counterparts \citep{earl_future_2015,bozarth_social_2017,hunt_horizontal_2023}. 

Social movements' understanding of their audiences can also shape their framing strategies \citep{snow_frame_1986,blee_social_2012}, and movements often customize their framing to appeal to different subsections of their audience \citep{andersen_islamic_2020,bergstrand_targeted_2022}. This is exemplified in online movements, where personalized content sharing allows for greater flexibility in adapting frames to mobilize diverse audiences, as opposed to relying on singular rigid collective action frames \citep{bennett_digital_2011,bennett_logic_2012}. On Twitter, posts are public by default and visible to one's full list of followers and beyond. However, people have diverse strategies to navigate this ``context collapse'' to reach intended sub-audiences \citep{marwick2011tweet}, such as through addressivity markers \citep{meraz_networked_2013}. Twitter's interactional features offer another mechanism for targeting intended audiences; different types of tweets (original ``broadcast'' tweets, quote tweets, and replies) have different intended audiences and communicative goals \citep{garimella2016quote}.



 \textbf{Issue Areas.}
We offer a brief background of major events in the United States from the time period we studied (2018-2019) that pertain to each issue area that we consider.

 \textbf{Guns.} On February 14, 2018, a gunman killed 17 people and wounded 17 others at Marjory Stoneman Douglas High School in Parkland, Florida \citep{aslett_what_2022}. In the wake of the tragic event, Parkland student survivors organized the March for Our Lives campaign, which rapidly gained national and international attention and reignited a longstanding political conflict over gun policy \citep{laschever_growth_2021}. March for Our Lives campaigned for and lobbied political leaders for increased gun control measures. Most notably, the student activists coordinated school-based, local, and national protests on a massive scale; the eponymous march in Washington D.C. and 764 other satellite locations drew 1.4 to 2.2 million participants, and the National Student Walkout on March 14, 2018 drew 1.1 to 1.7 million participants at 4,495 locations \citep{pressman_protests_2022}. The gun rights countermovement also mobilized during this time \citep{laschever_growth_2021}. 

Prior research has studied framing on Twitter and in news coverage following mass shootings, with a particular focus on partisan polarization \citep{demszky2019analyzing,lin_dynamics_2020,holody_neveragain_2022,zhang_reactive_2022}. Several articles have also compared framing strategies between gun control and gun rights SMOs \citep{steidley_framing_2017,merry_narrative_2018}. While gun control SMOs focus on child victims and mass shootings, gun rights SMOs focus on self-defense \citep{merry_narrative_2018}. Following the Parkland shooting, gun control SMOs further identified easy gun access as a problem and emphasized mobilization, while gun rights SMOs focused more on law enforcement's failures \citep{aslett_what_2022}. Focusing on student activists' diagnostic, prognostic, and motivational framing on Twitter, \cite{zoller_march_2022} finds that students identify lax gun policy as a problem, and frequently blame the NRA and their political influence. Prognostic framing includes promoting gun control legislation as a solution and boycotting/refusing NRA money as a tactic \citep{zoller_march_2022}. Surprisingly, gun control SMOs, including March for Our Lives, tend not to focus on race, even though gun violence primarily affects communities of color \citep{merry_narrative_2018,tergesen_marching_2021}

 \textbf{Immigration.} In May 2018, the U.S. Department of Justice under the Trump Administration implemented a ``zero tolerance'' policy at the U.S.-Mexico border, which required that all adult migrants who cross the border without permission be prosecuted while any children they crossed with were sent to separate detention facilities \citep{alamillo_framing_2019}. This represented a stark contrast to prior policy, in which families were kept together, either while awaiting their immigration cases or in deportation. In early June 2018, mainstream media outlets covered several individual cases of harm that this family separation policy caused.\footnote{\url{ https://www.splcenter.org/news/2022/03/23/family-separation-timeline}} In response, on June 15, the Department of Homeland Security publicly acknowledged that nearly 2,000 children had been separated from their families, and there was no clear plan for reunification. Media coverage, as well as public outrage, intensified when journalists and human rights advocates toured one of the children detention centers and reported on the poor conditions on June 17. After several days of escalating pressure, Trump signed an executive order on June 20 to end the family separation policy. In analyzing news coverage of the family separation crisis, \cite{alamillo_framing_2019} finds that most mainstream media outlets emphasized humanitarian concerns, particularly the harms that separations have on young children. Fox News, however, often talked about the risk of human trafficking at the border \citep{alamillo_framing_2019}. 

While media framing of immigration, immigrants, refugees, and asylum-seekers has been widely studied \citep{eberl_european_2018,seo_media_2022}, immigrant rights movements have only relatively recently been integrated into social movement scholarship \citep{mora_immigrant_2018}. Pro-immigration SMOs tailor their framing to different audiences in order to address ``four categories of aims: educating and persuading the general public, engaging non-supporters through dialogue, supporting and organizing migrants as activists, and building cooperative relationships with the authorities'' \citep{deturk_imagined_2023}. There has been related work on refugee discourse and activism through Twitter affordances \citep{siapera_refugees_2018,estrada_iamarefugee_2021}. For example, \citet{estrada_iamarefugee_2021} find that pro-refugee activists use the \#IAmARefugee hashtag on Twitter to express solidarity and engage in boundary work in opposition to Trump's ``Muslim Ban'', discursively constructing a moral ``us'' and an immoral ``them''. Finally, a small set of articles have analyzed diagnostic, prognostic, and motivational framing in the context of immigration, particularly among anti-immigration activists and the far-right \citep{dove2010framing,gagnon_far-right_2020,wahlstrom_dynamics_2021}. 


 \textbf{LGBTQ rights}. The contemporary LGBTQ rights movement has a long history in the U.S., with collective action frames emerging from the homophile movement in the 1950s and the civil rights protest wave in the 1960s \citep{valocchi_riding_2006}. Since the 1969 Stonewall uprising, Pride protests have been held annually in June and now occur in many cities in the U.S. and worldwide.\footnote{\url{https://www.loc.gov/lgbt-pride-month/about/}} We analyze Twitter data about LGBTQ rights from 2018-2019. This was several years after the 2015 Obergefell v. Hodges Supreme Court case that federally legalized same-sex marriage \citep{espinoza-kulick_multimethod_2020}, but before the massive wave of anti-trans legislation in the 2020s.\footnote{\url{https://translegislation.com/learn}} The Crowd Counting Consortium counted 3.7 million protest participants in June 2018, primarily from Pride events \citep{pressman_protests_2022}, with the majority of Pride participants attending demonstrations on June 24, 2018, when many major US cities including New York, San Francisco, Chicago, and Seattle, hosted parades.

LGBTQ activists and the Religious Right have been embroiled in a decades-long movement/countermovement dynamic that mutually participate in each others' framing processes \citep{liebler_queer_2009,stone_impact_2016}. As such, much prior work on framing with respect to LGBTQ rights has focused on comparisons between proponents and opponents of same-sex marriage in news coverage; while pro-LGBTQ activists emphasize civil rights and equality considerations, anti-LGBTQ activists frame same-sex marriage as a threat to Christian values and ``traditional'' heterosexual marriage \citep{hull_political_2001,warren_framing_2014}. Several papers have analyzed pro-LGBTQ and anti-LGBTQ dynamics on Twitter from the collective action perspective \citep{copeland_collective_2016,oktavianus_framing_2023}. For example, \citet{oktavianus_framing_2023} shows that in the context of the anti-LGBTQ \#UninstallGojek in Indonesia, anti-LGBTQ protestors primarily engaged with prognostic and motivational framing, while the pro-LGBTQ counterprotestors emphasized diagnostic considerations.


 \textbf{Computational Framing Analysis.}
There has been growing interest in automated frame detection to facilitate large-scale textual analysis. Computational approaches broadly fall into two camps: unsupervised and supervised methods, which parallel inductive and deductive coding in social sciences, respectively. Topic modeling is a popular unsupervised method, and has been used to analyze framing in news articles \citep{walter_news_2019}, social media posts \citep{tschirky_azovsteel_2023}, and online social movements such as Black Lives Matter, Me Too, climate advocacy, and the digitally-native Sleeping Giants movement \citep{li_beyond_2021,li_metoo_2021,chen_how_2022, klein_attention_2022}. Most closely aligned with our work, \cite{aslett_what_2022} use topic modeling to study frame contests between gun control and gun rights groups following the 2018 shooting at Marjory Stoneman Douglas High School in Parkland, Florida. They find that gun rights organizations emphasized the inefficacy of gun restrictions and highlighted law enforcement failure as the primary problem, while gun control groups identified easy access to guns as the main problem and emphasized mobilization. However, scholars argue that unsupervised approaches like topic modeling do not capture theoretically-grounded frames, in contrast to supervised classification with existing taxonomies \citep{nicholls_computational_2021, eisele_capturing_2023}.

Supervised frame detection involves first manually coding texts based on a pre-existing frame taxonomy, and then using this labeled data to train machine learning classification models. Prior work has implemented a wide variety of supervised classification models to detect frames, including support vector machines, random forest classifiers, neural networks, and fine-tuning pretrained language models such as RoBERTa \citep{khanehzar_modeling_2019,khanehzar_framing_2021,ali_survey_2022,eisele_capturing_2023}. A recent but quickly-growing body of literature is also exploring the potential of prompting large language models (e.g., ChatGPT) for both supervised and unsupervised frame analysis \citep{guo-etal-2022-capturing,mou_two_2022,roy-etal-2022-towards,ziems2023can}.

Especially within the field of NLP, much computational framing research uses the Policy Frames Codebook of issue-generic frames in U.S. news media \citep{boydstun2013identifying,boydstun2014tracking} and the associated Media Frames Corpus for training models \citep{card_media_2015}. In addition to news media, the Policy Frames Codebook has been used to analyze social media data, including politicians' tweets \citep{johnson_leveraging_2017}, public tweets about immigration \citep{mendelsohn2021modeling}, and online discussion posts \citep{hartmann_issue_2019}. The Policy Frames Codebook has also been deployed in analyzing news and social media data in non-English languages outside of the U.S. \citep{piskorski-etal-2023-multilingual,park2022challenges}. Other work has similarly focused on supervised frame detection of \cite{semetko2000framing}'s typology of issue-generic news frames \citep{burscher_teaching_2014,kroon_beyond_2022,alonso_del_barrio_framing_2023}. Although to a lesser extent, there has also been computational work considering other perspectives on framing, including entity-centric framing \citep{ziems-yang-2021-protect-serve,frermann_conflicts_2023} and morality framing \citep{roy-etal-2022-towards}. 

Several papers have developed corpora and computational models to identify issue-specific media frames for gun violence \citep{liu_detecting_2019,akyurek_multi-label_2020,tourni_detecting_2021} and immigration \citep{mendelsohn2021modeling}. We investigate these same issue areas but draw from a different theoretical framework focused on diagnostic, prognostic, and motivational core framing tasks for collective action mobilization \citep{benford_framing_2000}. Little prior work has focused on automatically classifying these core framing tasks, with several notable exceptions. \cite{hsu_frame_2016} use supervised machine learning to classify diagnostic, prognostic, and motivational framing in messages from the Taiwanese anti-curriculum student movement. \cite{alashri_climate_2016} and \cite{fernandez-zubieta_digital_2022} build supervised models to detect these core framing tasks in news articles and social media posts about climate change and climate action. Social movement scholars recognize the need for empirical, large-N, and comparative framing research in addition to in-depth interpretive case studies \citep{snow_emergence_2014}, but such endeavors have been limited by the relative lack of sociologically-grounded computational approaches. Our work addresses this methodological gap and showcases the utility of computational methods for such empirical analyses. 

\section{Data}

We now turn to our data collection, codebook creation, and annotation procedures. We first manually annotate framing strategies in a random sample of data in order to train models to automatically infer labels for the entire dataset of nearly two million tweets.

 \textbf{Data Collection.}
We study the framing of online social movements with data from \citet{bozarth_keyword_2022}, which contains tweets from the Twitter Decahose 10\% sample from 2018-2019. Using collective action event, participation, and issue area data from the Crowd Counting Consortium (CCC),\footnote{\url{https://github.com/nonviolent-action-lab/crowd-counting-consortium}} \citet{bozarth_keyword_2022} select five issue areas that have the most variance in the number and size of their associated collective action events: \textit{government}, \textit{healthcare}, \textit{LGBTQ rights}, \textit{guns}, and \textit{immigration}. They collect data from two months for each issue: one month characterized by high protest activity and one month with average levels of protest activity. We then select a subset of this data that covers three issue areas: \textit{guns}, \textit{immigration}, and \textit{LGBTQ rights} for two reasons. First, these three issue areas have one overlapping month of data (June 2018), which facilitates direct comparisons. Second, we manually inspect the collective action event claims (another field in the CCC data) and qualitatively observe that events for these three issue areas had more cohesive and unified goals, and are thus more closely aligned with our conceptualization of social movements.

For each issue area, \citet{bozarth_keyword_2022} develop lexicons for keyword-based tweet collection by combining and validating a set of machine learning and embedding-based keyword expansion techniques, starting with a seed set of frequent hashtags posted during collective action events. Note that \citet{bozarth_keyword_2022} do not distinguish between progressive and conservative movement activities, and thus opposing movements surrounding the same issues are represented. For example, both \textit{guncontrolnow} and \textit{guncontrolnever} are keywords for the \textit{guns} issue area. The resulting dataset contains 1.85M tweets in total, with 822K tweets about \textit{guns}, 763K tweets about \textit{immigration}, and 268K tweets about \textit{LGBTQ rights}. The dataset includes replies and quote tweets, but excludes retweets with no additional commentary. The counts by month are shown in Table \ref{tab:monthly-counts}. Major actions during high activity months include Pride parades for \textit{LGBTQ rights}, protests against family separation at the US-Mexico border for \textit{immigration}, and youth-led demonstrations following the school shooting in Parkland, Florida for \textit{guns}. 

We opt to use this dataset as it provides several distinctive advantages for social movement framing analysis. Twitter has long been recognized as a primary site for social movement activism \citep{meraz_networked_2013}; the high volume of relevant messages on Twitter facilitates both rich cross-sectional and fine-grained temporal analyses of framing strategies. The metadata linked to tweets, such as the exact time of posting and author information, further enable such analyses. In contrast with many other datasets that focus on a single social movement, \citet{bozarth_keyword_2022}'s dataset is particularly valuable because it covers multiple issues with active associated movements in the same time period. Moreover, each issue area encompasses both progressive and conservative movements, providing the opportunity for comparative content analysis across both issue and ideological stances. With data from months with both high and average levels of offline collective action activity, \citet{bozarth_keyword_2022}'s dataset can help us understand the relationship between online discourse and offline events. Finally, this dataset was collected with a validated keyword expansion method that retrieves a much broader set of tweets than those collected via single keywords or hashtags.  

At the same time, this dataset presents several limitations that can be addressed in future work. While the dataset includes months characterized by different levels of activity for each issue area, it still encompasses a relatively small stretch of time from 2018-2019. It is also limited to one social media platform that may not necessarily be representative of all social movement communication. The inclusion of multiple issue areas is a key advantage of this dataset, but there are many issue areas that are not considered, particularly outside of the US context. The dataset and our methodology exclusively focus on textual communication, and do not include modalities such as audio, video, or images. While out of scope for the present work, future research ought to investigate if the findings from our data generalize to other issues, platforms, time periods, languages, modalities, and cultural contexts.

\begin{table}[htbp!]
\centering
\resizebox{.7\textwidth}{!}{%
\begin{tabular}{@{}cccc@{}}
\toprule
\textbf{Issue} & \textbf{Protest Activity} & \textbf{Month} & \textbf{Tweet Count} \\ \midrule
guns           & high                      & March 2018     & 633,027              \\
guns           & average                   & June 2018      & 189,134              \\ \midrule
immigration    & high                      & June 2018      & 513,284              \\
immigration    & average                   & July 2018      & 249,776              \\ \midrule
LGBTQ rights   & high                      & June 2018      & 172,006              \\
LGBTQ rights   & average                   & April 2019     & 95,695               \\ \bottomrule
\end{tabular}%
}
\caption{Monthly tweet counts by issue and protest activity level}
\label{tab:monthly-counts}
\end{table}

\begin{table}[htbp!]
\centering
\resizebox{\textwidth}{!}{%
\begin{tabular}{@{}ccl@{}} \toprule
\textbf{Category} &
  \textbf{Sub-Category} &
  \textbf{Brief Codebook Description} \\ \toprule

Stance &
  \cellcolor[HTML]{EFEFEF} Stance &
  \cellcolor[HTML]{EFEFEF}\begin{tabular}[c]{@{}l@{}}Based on the text, would you guess that this message was \\ written by someone with a progressive, conservative, or \\ neutral/unclear attitude towards the specified issue?\end{tabular} \\ \midrule
\multirow{2}{*}{Diagnostic} &
  Identification &
  \begin{tabular}[c]{@{}l@{}}Does this message identify a social or political problem?\\ \textit{Ex: homophobia, school shootings, family separation at the border}\end{tabular} \\ 
 &
  \cellcolor[HTML]{EFEFEF} Blame &
  \cellcolor[HTML]{EFEFEF}\begin{tabular}[c]{@{}l@{}}Does this message assign blame for a societal problem?\\ \textit{Ex: to the government, corporations, socioeconomic systems}\end{tabular} \\ \midrule
\multirow{4}{*}{Prognostic} &
  Solutions &
  \begin{tabular}[c]{@{}l@{}}Does this message propose solutions for a societal problem?\\ \textit{Ex: changes in policies, political leaders, or societal norms}\end{tabular} \\
 &
  \cellcolor[HTML]{EFEFEF} Tactics &
  \cellcolor[HTML]{EFEFEF}\begin{tabular}[c]{@{}l@{}}Does this message discuss strategies or tactics for\\ achieving a movement's goals?\\\textit{Ex: protests, boycotts, petitions, contacting politicians}\end{tabular} \\
 &
  Solidarity &
  \begin{tabular}[c]{@{}l@{}}Does this message express support or solidarity for a movement?\\ \textit{Ex: celebrating a movement, honoring activists, raising visibilit}y\end{tabular} \\
 &
  \cellcolor[HTML]{EFEFEF} Counterframing &
  \cellcolor[HTML]{EFEFEF}\begin{tabular}[c]{@{}l@{}}Does this message explicitly challenge arguments \\ made by the opposing side?\end{tabular} \\ \midrule
Motivational &
  Motivational &
  \begin{tabular}[c]{@{}l@{}}Does this message try to convince readers to join, \\ participate in, or support a social movement through calls to action?\end{tabular} \\ \bottomrule
\end{tabular}%
}
\caption{Annotation typology and codebook descriptions for stance, core framing tasks, and frame elements}
\label{tab:typology}
\end{table}

\begin{figure}[htbp!]
\centering
\includegraphics[width=.7\textwidth]{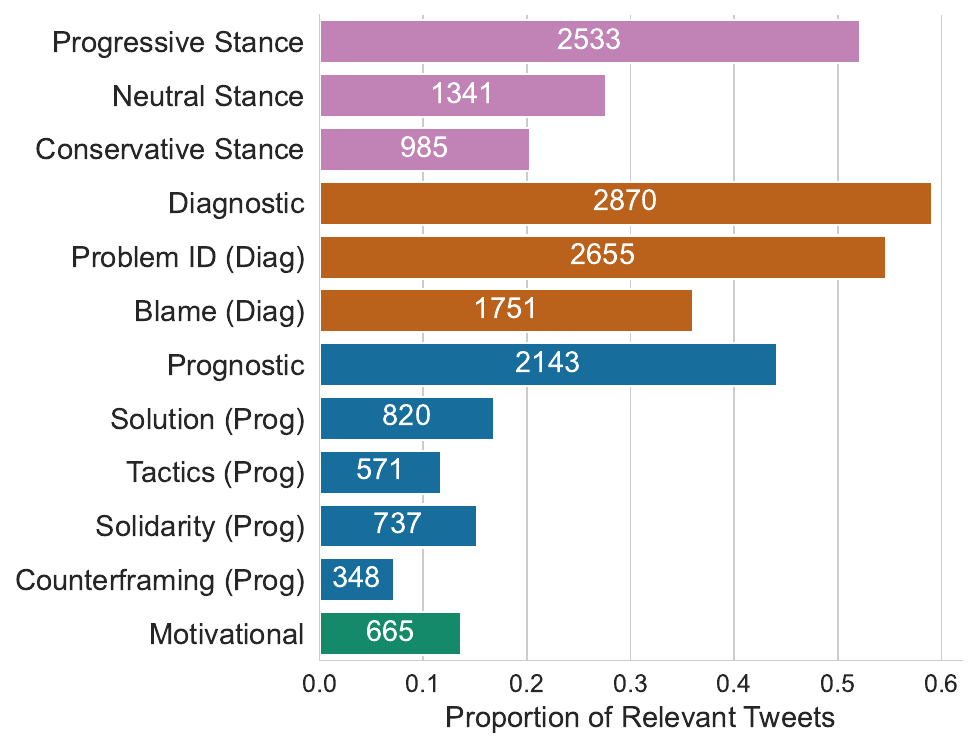}
\caption{Label prevalence in the annotated dataset.}
 \floatfoot{ \RaggedRight \normalsize \textit{Note.} As a proportion of relevant tweets. Raw counts are shown inside the bars.}
\label{fig:label_frequency}
\end{figure}

\section{Annotation}
We developed and iteratively refined a codebook based on theoretical definitions \citep{benford_framing_2000,della_porta_social_2006} and existing codebooks for characterizing diagnostic, prognostic, and motivational framing strategies \citep{goh_protesting_2016,phadke_framing_2018,phadke_many_2020}. 

 \textbf{Framing Typology.}
Because Boolean keyword search by itself is a coarse proxy for relevance, we follow \citet{bozarth_keyword_2022}'s suggestion to first categorize tweets as \textit{relevant or irrelevant} to an issue area. We further code relevant tweets for \textit{stance} and \textit{diagnostic}, \textit{prognostic}, and \textit{motivational} framing strategies (Table \ref{tab:typology}). Stance can take one of three mutually-exclusive values: progressive (pro-gun control, pro-immigration, pro-LGBTQ), conservative (pro-gun rights, anti-immigration, anti-LGBTQ), or neutral/unclear from the text. Guided by \citet{benford_framing_2000} and \citet{goh_protesting_2016}, we operationalize and code for diagnostic framing based on the presence of any of two sub-categories: \textit{problem identification} and \textit{blame attribution}. Similarly, we code for prognostic framing based on the presence of any of four sub-categories based in \citet{benford_framing_2000}'s definition: proposing \textit{solutions}, discussing movement \textit{tactics}, expressing \textit{solidarity}, and engaging in \textit{counterframing}. Throughout this paper, we refer to the broader framing categories as \textbf{core framing tasks}, and the narrower sub-categories as \textbf{frame elements}. Note that motivational framing was directly coded, and is thus considered both a core framing task and a frame element. 

\begin{table}[htbp!]
\centering
\caption{Label prevalence in the annotated dataset by issue area.}
\label{tab:label_frequency_by_issue}
\resizebox{.5\textwidth}{!}{%
\begin{tabular}{@{}lccc@{}}
\toprule
 & Guns & Immigration & LGBTQ \\ \midrule
Progressive Stance & 566 & 627 & 1340 \\
Neutral Stance & 357 & 477 & 507 \\
Conservative Stance & 362 & 519 & 104 \\ \midrule
Diagnostic & 761 & 1172 & 937 \\
Problem ID (Diag) & 661 & 1079 & 915 \\
Blame (Diag) & 517 & 837 & 397 \\ \midrule
Prognostic & 644 & 689 & 810 \\
Solution (Prog) & 258 & 435 & 127 \\
Tactics (Prog) & 256 & 141 & 174 \\
Solidarity (Prog) & 112 & 83 & 542 \\
Counterframing (Prog) & 154 & 123 & 71 \\ \midrule
Motivational & 220 & 208 & 237 \\ \bottomrule
\end{tabular}
}
\end{table}

Starting with a preliminary codebook, the first author and two undergraduate research assistants completed a pilot annotation task of 150 tweets. From the resulting discussions, we adapted the codebook for Twitter data, expanded the neutral stance code to include tweets with unclear stance, and identified the need for a separate \textit{solidarity} frame element \citep{hon_social_2016}. The first author and one undergraduate research assistant proceeded to conduct several iterations of annotator training. Each iteration consisted of 150 tweets, evenly split across issues and protest activity periods. After independent labeling, annotators met to resolve all disagreements and agree on any minor clarifications and modifications to the codebook. Inter-annotator agreement (Krippendorff's $\alpha$) was calculated after each round and sufficient agreement was reached after five rounds, with $\alpha \geq 0.75$ for all categories except for \textit{counterframing} for which $\alpha=0.66$. Following the fifth round, annotators proceeded independently until a total of 6,000 tweets were labeled (equally distributed across issues and protest activity level periods). 4,859 (81\%) of these tweets were coded as \textit{relevant} and further coded for stance and framing.

The overall presence of each stance and framing category in our manually-labeled dataset is shown in Figure \ref{fig:label_frequency}, and separated by issue area in Table \ref{tab:label_frequency_by_issue}. \textit{Progressive} stance and \textit{diagnostic} framing are the most frequent categories overall, while \textit{counterframing}, \textit{tactics}, and \textit{motivational} framing are the least frequent categories. However, this varies across issues. For example, \textit{solidarity} is much more frequent for LGBTQ-related tweets than for other issues, while \textit{solution} is less frequent. Tweets may cue anywhere from zero core framing tasks to all three (Table \ref{tab:frame_count_by_issue}. The plurality of relevant tweets in all issue areas engages in one core framing task. About 19\% of relevant tweets are not labeled with any collective action framing strategies. While these tweets are relevant to issue areas, they tend to not be as relevant to particular social movements. Many tweets with no frame labels are jokes, posts about author's typical everyday experiences, or short direct responses to unobserved tweets or external links with little additional information.   

\begin{table}[htbp!]
\centering
\caption{Number of core framing tasks in tweets.}
\label{tab:frame_count_by_issue}
\resizebox{.5\textwidth}{!}{%
\begin{tabular}{@{}ccccc@{}}
\toprule
Frame Count& Guns& Immigration& LGBTQ& Total\\ \midrule
0 & 223 & 244 & 444 & 911 \\
1 & 618 & 810 & 1091 & 2519 \\
2 & 325 & 448 & 355 & 1128 \\
3 & 119 & 121 & 61 & 301 \\ \midrule
Average& 1.26 & 1.27 & 1.02 & 1.17 \\ \bottomrule
\end{tabular}
}

\vspace{0.5em}
{\RaggedRight \textit{Note.} Number of tweets containing zero, one, two, or three core framing tasks, and the average number of core framing tasks in each tweet separated by issue area.}
\end{table}

\section{Classifying Framing Strategies}

We operationalize our taxonomy as a set of four classification problems. First is binary \textbf{relevance} classification. Second is a 3-class \textbf{stance} classification, where the \textit{progressive}, \textit{conservative}, and \textit{neutral/unclear} outputs are mutually exclusive. Third is \textbf{core framing task} classification. Because a tweet may be labeled with none, some, or all three of the \textit{diagnostic}, \textit{prognostic}, and \textit{motivational} strategies, we treat this as a binary, multi-label problem. Fourth is \textbf{frame element} classification, which includes the seven frame elements that were directly coded for (\textit{problem identification}, \textit{blame}, \textit{solution}, \textit{tactics}, \textit{solidarity}, \textit{counterframing}, and \textit{motivational}). This is similarly a binary multi-label classification problem where anywhere from 0-7 frame elements may be present in a tweet.

 \textbf{Model Setup.} For each of these four classification problems, our goal is to train computational models to predict the appropriate labels. We use a common approach in NLP, where we build our classifiers on top of RoBERTa \citep{liu_roberta_2019}, a highly-parameterized pretrained language model, which has been trained on vast amounts of unlabeled data. Following standard practice, we first finetune the parameters of the RoBERTa-base model on the full corpus of 1.85M tweets in order to adapt the language model to better recognize linguistic patterns in tweets related to social movements.\footnote{All RoBERTa-based models were trained using the \texttt{simpletransformers} package. We finetuned RoBERTa for five epochs with a batch size of 128, with all other hyperparameters set to the package defaults. Using two NVIDIA V100 GPUs, finetuning took slightly under three hours.} We then separately train the finetuned RoBERTa model for each of the four classification tasks. We split our labeled sample into training and testing splits, containing 80\% and 20\% of the data, respectively. Within the training set, we use 5-fold cross-validation to refine and compare models. We train separate models for each issue, as well as a combined model that includes data from all issues in training and evaluation.\footnote{We trained each classifier on one GPU for 20 epochs and a batch size of 32. Each model was trained in under 30 minutes.} Given that we expect framing strategies to somewhat generalize across issues, we first consider whether to treat issues together or separately. As part of our preliminary investigation, we first compare training separate models for each issue, as opposed to pooling data from all issues, and using that combined data for training models that do not explicitly distinguish between issues. 


 \textbf{Model Evaluation.}
Based on preliminary experiments, with results in Table \ref{tab:combined_vs_specific}, combined-issue models trained on pooled data across issues outperform issue-specific models. We thus decide to proceed with the combined-issue models for further evaluation and analysis. The higher test set performance is likely due to our cross-validation setup. Each development model was only trained on 4/5 folds of the training set (64\% of the total dataset). Models evaluated on the test set were trained on the full training set (80\% of the total dataset).\footnote{Note that we do not deduplicate our data because our primary goal is to make accurate predictions on the full dataset. As such, these numbers may give an artificially high sense of model performance, because there will be either identical or near-identical tweets that exist in both training and testing data. We partially mitigate this problem by excluding retweets (but not quote tweets) from our dataset. Although further deduplication could be used to get more precise estimates of performance on truly unseen data, this would be both difficult to accomplish effectively (because of near-duplicate tweets) and would give less meaningful estimates for our goals.}

\begin{table}[htbp!]
\centering
\caption{Average F1 scores for combined-issue and issue-specific models.}
\vspace{-1em}
\resizebox{.8\textwidth}{!}{%
\begin{tabular}{@{}llccc@{}} \\ \toprule
      &                & \multicolumn{3}{c}{Evaluation Issue}          \\
Split & Training Data  & Guns          & Immigration   & LGBTQ         \\ \toprule
\multirow{2}{*}{Development} & All Issues & \textbf{0.710 (0.121)} & \textbf{0.697 (0.177)} & \textbf{0.673 (0.234)} \\
      & Single Issue & 0.680 (0.145) & 0.651 (0.223) & 0.634 (0.267) \\  \midrule
\multirow{2}{*}{Test}       & All Issues & \textbf{0.734 (0.121)} & \textbf{0.709 (0.153)} & \textbf{0.694 (0.204)} \\
      & Single Issue & 0.706 (0.117) & 0.679 (0.218) & 0.643 (0.248) \\ \bottomrule

\end{tabular}%
}

\vspace{0.5em}
{\RaggedRight \textit{Note.} The scores shown are averages of F1 scores per label. Standard deviations are in parentheses. Development scores are averaged over five cross-validation folds.}
\label{tab:combined_vs_specific}
\end{table}

\begin{table}[htbp!]
\centering
\resizebox{.5\textwidth}{!}{%
\begin{tabular}{@{}cccc@{}}
\toprule
\textbf{Issue} & \textbf{Liberal} & \textbf{Neutral} & \textbf{Conservative} \\ \midrule
Guns           & 0.824            & 0.559                    & 0.652                 \\
Immigration    & 0.763            & 0.582                    & 0.798                 \\
LGBTQ          & 0.879            & 0.656                    & 0.410                 \\ \bottomrule
\end{tabular}%
}
\caption{Stance classification F1-scores by issue, evaluated on the test.}
\label{tab:stance_test}
\end{table}

We observe high performance of the combined-issue relevance classifier, with a test F1 score of 0.968 (precision = 0.959, recall = 0.976). Table \ref{tab:stance_test} (and Appendix Table \ref{tab:stance_dev}) show per-issue stance classification F1-scores. Likely due to the skew towards \textit{liberal} tweets in the training data, performance is overall highest for \textit{liberal} tweets, though the stance model has high performance for identifying \textit{conservative} tweets about immigration, which is the issue area with the highest frequency of conservative tweets in the labeled dataset (Table \ref{tab:label_frequency_by_issue}). The lowest F1-scores occur for identifying \textit{conservative} tweets about LGBTQ rights, which we attribute again to a very imbalanced dataset with few anti-LGBTQ messages.

Table \ref{tab:frame_f1} contains per-label, macro, and micro F1 scores for the \textit{core framing task} and the \textit{frame element} classifiers. Both models perform reasonably well, with the \textit{core framing task} model achieving a micro-F1 score of 0.815 and the \textit{frame element} model achieving a micro-F1 score of 0.763. As these results are substantially better than prior computational frame analysis work \citep{mendelsohn2021modeling,park2022challenges}, we consider our models to be sufficient overall for inferring framing strategies in the full corpus, and for conducting analysis with the inferred labels. 

\begin{table}[htbp!]
\centering
\resizebox{.45\textwidth}{!}{%
\begin{tabular}{@{}lrr@{}}
\toprule
Core Framing Task & Dev F1 & Test F1 \\ \midrule
Diagnostic        & 0.880                       & 0.885                       \\
Prognostic        & 0.761                      & 0.765                       \\
Motivational      & 0.657                      & 0.690                        \\ \midrule
Macro F1          & 0.766                      & 0.780                        \\
Micro F1          & 0.811                      & 0.815                       \\ \bottomrule
\end{tabular}%
}
\hfill
\resizebox{.45\textwidth}{!}{%
\begin{tabular}{@{}lrr@{}}
\toprule
Frame Elements        & Dev F1 & Test F1 \\ \midrule
Problem ID (Diag)     & 0.856                      & 0.869                       \\
Blame (Diag)          & 0.769                      & 0.773                       \\
Solution (Prog)       & 0.703                      & 0.685                       \\
Tactics (Prog)        & 0.617                      & 0.594                       \\
Solidarity (Prog)     & 0.773                      & 0.777                       \\
Counterframing (Prog) & 0.398                      & 0.473                       \\
Motivational          & 0.662                      & 0.697                       \\ \midrule
Macro F1              & 0.683                      & 0.695                       \\
Micro F1              & 0.758                      & 0.763                       \\ \bottomrule
\end{tabular}%
}

\caption{F1 scores for core framing tasks and frame elements.}
\label{tab:frame_f1}
\end{table}

At the same time, we observe variability in model performance across individual labels; some categories with lower performance are quite rare in the data. For example, the \textit{core framing task} classifier has the lowest F1-score for \textit{motivational} framing, which only appears in 14\% of relevant labeled tweets (Figure \ref{fig:label_frequency}), while 59\% of relevant tweets contain \textit{diagnostic} framing, the highest category. Indeed, F1 scores and support (number of labels in the evaluation set) are highly correlated, with Pearson correlation of 0.84 and 0.82 for all categories in the development and test sets, respectively. 

When taking the imbalanced sample into account, model performance is even more encouraging. For example, \textit{tactics} has a seemingly low F1-score of 0.594, but a random baseline classifier by comparison has in expectation an F1-score of just 0.191. Nevertheless, we decide to exclude stance and framing categories with F1-scores below 0.5 from further analysis in order to increase the reliability of our analysis. We thus omit the lowest performing frame element, \textit{counterframing}. For all other categories, we infer labels on the full corpus. We first identify 1.48M (out of 1.85M) \textit{relevant}, among which we identify \textit{stance}, \textit{core framing tasks}, and \textit{frame elements}.

\section{Linguistic properties of core framing tasks}

In our data annotation and classification tasks, we deconstruct social movement messages into core framing tasks and frame elements. These categories are still higher-order constructs which can themselves be decomposed into lower-level units, namely linguistic features. Inspired by calls for more framing research to focus on such micro-level processes of meaning construction within texts \citep{johnston_methodology_1995,hedley_microlevel_2007,vicari_measuring_2010}, we explore the linguistic features used to accomplish each core framing task. Beyond establishing that different core framing tasks are characterized by different linguistic markers, this analysis is primarily intended to give us a richer understanding of \textit{how} authors communicate each core framing task, and thus offers a bridge between micro-level discourse analysis and higher-order content analysis approaches to social movement texts. From a computational perspective, such fine-grained linguistic analysis is also advantageous, as it offers a way to gain insights into our large-scale dataset and machine learning models. Linguistic features associated with messages containing each core framing task are also likely to be the same signals on which our models place high predictive weight in classification. As interpretability with large language models such as RoBERTa remains an open challenge, such linguistic analyses can provide insights into the models' decision-making process.

We identify linguistic features associated with each core framing task by calculating the log-odds ratio with informative Dirichlet prior within each issue area \citep{monroe2008fightin}. This metric identifies features that are statistically overrepresented in one corpus relative to another. For example, we compare the frequencies of linguistic features in \textit{diagnostic} tweets about \textit{immigration} to frequencies in all tweets about \textit{immigration}. We calculate log-odds statistics for five linguistic features: words, verbs, adjectives, subject-verb pairs, and verb-object pairs. Our consideration of subject, verb, and object relations is inspired by \citet{johnston_methodology_2013}, who argues that these structures are meaningful units for social movement frame analysis. We use the \texttt{en-core-web-sm} model in the SpaCy Python package for tokenization, part-of-speech tagging, and dependency parsing to extract subject-verb-object structures from tweets. For all features except words, we also preprocess features by lemmatizing (normalizing different morphological forms of the same words) with SpaCy. After calculating log-odds, we select the top 15 of each feature most associated with each core framing task within each issue area for qualitative analysis. We identify several themes that we will focus on for the remainder of this section, but the full log-odds results can be found in Appendix Table \ref{tab:logodds}.

Patterns in \textbf{pronoun person marking} in the log-odds results suggest boundary framing processes that clearly identify protagonists and antagonists in conflict, literally ``us'' vs. ``them'' \citep{snow_frame_1986, vicari_measuring_2010, das_framing_2021}. For all issue areas, 3rd person pronouns (e.g., \textit{they, their, he} appear in the top 15 words most associated with diagnostic framing. Words most associated with prognostic framing include 1st person pronouns (e.g., \textit{we, our}) and subject-verb tuples across issues include phrases such as \textit{we need}, \textit{we want}, and \textit{we have}. Finally, 2nd person pronouns (e.g., \textit{you}, \textit{your}) are among the words most associated with diagnostic framing within immigration and LGBTQ tweets (while 2nd person pronouns are still significantly associated with motivational framing for gun tweets, it is crowded out by words from tweets motivating people to participate in gun giveaways and auctions). We further corroborate this relationship between pronoun person and framing with logistic regression models, which shows that diagnostic framing has the strongest positive association with 3rd person pronouns, prognostic framing is most associated with 1st person pronouns, and motivational framing is most associated with 2nd person pronouns (see Appendix Figure \ref{fig:pronouns} for more details)

Qualitative analysis of the top 15 adjectives and verbs reveals the centrality of \textbf{moral language} in both diagnostic and prognostic framing. Across all issues, there are several adjectives most associated with diagnostic framing that express moral disapproval or disgust: \textit{bad}, \textit{sick}, \textit{disgusting} for guns; \textit{inhumane}, \textit{cruel}, \textit{evil}, \textit{sick}, \textit{wrong}, \textit{disgusting} for immigration; \textit{bad}, \textit{wrong}, \textit{disgusting}, \textit{hateful} for LGBTQ. On the other end, top verbs associated with prognostic framing for all issues include deontic modal verbs such as \textit{need}, \textit{should}, and \textit{must}, which often signal moral obligation \citep{vicari_measuring_2010}. 

Closer analysis of verbs and their subject and object arguments provide additional insight into how movements deploy each core framing task, and commonalities across issue areas. For example, top verbs associated with diagnostic framing, such as \textit{kill}, \textit{attack}, \textit{destroy}, and \textit{murder}, suggest that violence is a commonly-identified problem. Perhaps less obviously, neglect is a commonly-identified problem across all issue areas, with verbs such as \textit{fail}, \textit{ignore}, \textit{lie}, \textit{refuse}, and \textit{deny}. Messages from all issue areas engage in motivational framing by emphasizing the necessity of action (\textit{need}, \textit{do\_something}, encouraging readers to join a movement (\textit{join}, \textit{join\_today}, \textit{join\_us}), and encouraging readers to pass or support legislation (\textit{pass\_legislation}, \textit{support\_bill}, \textit{tell\_congress}). 

For all three issues, many of the same features are associated with each core framing task, suggesting a degree of cross-issue stability and generalizability in how core framing tasks are linguistically constructed. However, analysis of individual words and verb-object pairs in particular reveal issue-specific components of core framing tasks. For example, Table \ref{tab:logodds} shows that some linguistic features are associated with diagnostic framing only within the \textit{guns} issue area, such as \textit{school}, \textit{shooting}, \textit{blame\_nra}, and \textit{kill\_child}, while features associated with diagnostic framing for \textit{LGBTQ rights} includes \textit{homophobia}, \textit{homophobic}, and \textit{use\_slur}. Similarly, features associated with prognostic framing include \textit{ban\_weapon} and \textit{stop\_violence} for \textit{guns}, but \textit{celebrate\_pride} and \textit{raise\_awareness} for \textit{LGBTQ rights}.

These findings highlight that log-odds ratios enable us to compare and contrast how the same core framing tasks are constructed across issue areas. These methods can be further applied to compare linguistic features of frames across axes beyond the issue area, such as ideological stance, time periods, and authors' social or professional identities. While largely beyond the scope of this work, we offer two brief examples. First, top features for diagnostic framing in progressive immigration tweets refer to the Trump administration family separation policy, such as \textit{\#trumpconcentrationcamps}, \textit{\#wherearethechildren}, and \textit{separate\_child}. On the other hand, top features for diagnostic framing in conservative immigration tweets emphasize criminal activity, such as \textit{illegally}, \textit{break\_law}, \textit{commit\_crime}, and \textit{illegal\_vote}. Second, language associated with prognostic framing in progressive gun-related tweets largely focuses on the March for Our Lives demonstrations and emphasis on change (\textit{marchforourlives}, \textit{make\_change}, \textit{find\_march}), while terms associated with prognostic framing in conservative gun-related tweets largely focus on preserving 2nd Amendment rights (\textit{defendthesecond}, \textit{protect\_right}, \textit{defend\_2a}).




\section{Frame variation across sociocultural contexts}

As social movements are situated within a broader time and space, their framing can shape--and be shaped by--social and cultural context \citep{snow_frame_1986,benford_framing_2000}. We address how framing varies across such contextual factors. We select three movement-level (issue area, stance, and protest activity period) and two message-level (author role and tweet type) dimensions for this analysis. We will first discuss our operationalization of each sociocultural factor, followed by Table \ref{tab:sociocultural-questions}, which summarizes each factor and its corresponding research question.


\textbf{Issue Area}. To broadly understand how social movement framing strategies vary across different contexts, we compare attention to each core framing task across the three issue areas included in our corpus: \textit{guns}, \textit{immigration}, and \textit{LGBTQ rights}. We may expect to see higher rates of prognostic and motivational framing for \textit{LGBTQ rights} because of the large number and size of Pride events in June. However, prognostic framing could also be more common for the other issue areas, as people may advocate for specific and concrete policy solutions for \textit{guns} and \textit{immigration}. While there have been long-sustained movements in all three issue areas, there were major offline events related to \textit{guns} and \textit{immigration} in this time period (the Parkland shooting and Trump's immigration policy, respectively) that could have implications for framing; for example, diagnostic framing could be most common for \textit{immigration} if messages primarily express outrage at and criticisms of family separation. It is important to note that differences in framing strategies across issue areas may not necessarily be consequences of inherent properties of issues, as variation across issue areas may reflect differences in the nature of events and activities being discussed.

\textbf{Stance.}  Although it may not necessarily be the case that each combination of stance and issue (e.g., \textit{progressive} tweets about \textit{immigration}) constitute a single social movement, comparing \textit{progressive}, \textit{conservative}, and \textit{neutral/unclear} message framing could help us better understand the nature of movement/countermovement relationships. We expect both \textit{progressive} and \textit{conservative} tweets to be more likely to contain collective action frames compared to \textit{neutral/unclear} tweets. However, it is not clear if there would be differences between \textit{progressive} and \textit{conservative} tweets in engagement with each core framing task.

\textbf{Protest activity.} We assess how attention to core framing tasks on Twitter varies between months characterized by \textit{high protest activity} and \textit{average protest activity} for each issue area. We expect that during periods of high protest activity, people would use Twitter to coordinate offline protests, discuss other tactics, and motivate people to participate, thus contributing to higher rates of prognostic and motivational framing during those periods. Note that while our dataset spans many different kinds of protest activities, ranging from Pride celebrations to school walkouts and marches, this analysis does not account for such differences in the nature of protest activity.

\textbf{Author role.} We compare framing across different author roles: \textit{journalist}, \textit{SMO}, or \textit{other} (mostly members of the public). Prior work has identified similarities in journalists' and activists' diagnostic frames on Twitter in the context of the 2014 Ferguson protests, but found that only activists tweet about protest action plans \citep{barnard_tweeting_2018}. In their study of \#MeToo, \citet{xiong_hashtag_2019} argue that many hashtags used by SMOs are action-oriented, event-specific, and highlight specific activists. We thus may expect SMOs to engage in more prognostic and motivational framing. At the same time, many hashtags were also used for problem identification \citep{xiong_hashtag_2019}, which could lead us to anticipate higher rates of diagnostic framing among SMOs compared to the other groups. 

We identify journalists based on two existing lists of Twitter handles of journalists associated with major U.S. outlets \citep{tauberg2022top,gotfredsen2023journalists}.\footnote{Data available at \url{https://github.com/taubergm/Top10000Journalists/tree/main}\\and \url{https://github.com/TowCenter/journalists-twitter-activity}} We follow \citet{bozarth_keyword_2022}'s procedure to identify SMO accounts. Specifically, the Crowd Counting Consortium data for most protest events includes the name of the SMO that organized the event. We then use the Twitter Search API to retrieve likely Twitter account matches for organizers of all protest events between 2017-2019 in any of the three issue areas. Out of 1.48M tweets classified as relevant, 4,218 are labeled with the \textit{journalist} author role and 5,817 with the \textit{SMO} author role. The remaining tweets are labeled with the \textit{other} author role. Note that this mostly consists of tweets from the general public, but also includes content from citizen activists and politically-motivated users \citep{terechshenko2020influential}.

\textbf{Tweet type.} We compare framing across different types of interactions on Twitter: \textit{broadcasts} (original tweets), \textit{quote tweets} (retweet with additional commentary), and \textit{replies}. In contrast to \textit{broadcasts}, both \textit{replies} and \textit{quote tweets} are responses to other messages. However, they differ in intended audience and communicative goals \citep{garimella2016quote}; while \textit{replies} are intended for the original post author and people engaged with the original post, \textit{quote tweets} add commentary or additional context typically for a broader audience, including the quote tweeter's own network of followers. Prior work on discussions in a Swedish far-right Facebook group suggests that broadcasts may be more likely to identify problems, while replies may be more likely to assign blame and propose solutions \citep{wahlstrom_dynamics_2021}; however, it is not clear if such patterns would generalize to our context of US-centered social movements on Twitter.

\textbf{Model.} We fit logistic regression models to analyze the relationship between framing and these five sociocultural factors. Sociocultural factors are included as independent variables, with the following reference levels: \textit{guns} for issue area, \textit{neutral/unclear} for stance, \textit{average} for protest activity level, \textit{other/public} for author role, and \textit{broadcast} for tweet type. Because our stance labels come from classifiers with imperfect accuracy (and especially low performance for identifying anti-LGBTQ tweets), we additionally fit models that exclude stance. We model eight dependent variables: three core framing tasks (\textit{diagnostic}, \textit{prognostic}, and \textit{motivational}) and five frame elements (\textit{problem identification}, \textit{blame}, \textit{solution}, \textit{tactics}, and \textit{solidarity}). Each dependent variable is represented as a binary indicator representing the presence or absence of that framing strategy for a given message. 

\begin{table}[htbp!]
\centering
\resizebox{\textwidth}{!}{%
\begin{tabular}{@{}ll@{}}
\toprule
Factor           & Question                                                                               \\ \midrule
\rowcolor[HTML]{EFEFEF} 
Issue Area       & How does framing vary across three issue areas: immigration, guns, and LGBTQ rights?   \\
Stance           & How does framing vary between progressive, conservative, and neutral/unclear messages? \\
\rowcolor[HTML]{EFEFEF} 
Protest activity & How does framing vary across tweets from high and average protest activity months?     \\
Author role &
  \begin{tabular}[c]{@{}l@{}}How does framing vary between journalists, social movement organizations (SMOs), \\ and others (neither SMOs nor journalists)?\end{tabular} \\
\rowcolor[HTML]{EFEFEF} 
Tweet type &
  \begin{tabular}[c]{@{}l@{}}How does framing vary across different types of Twitter interactions: \\ broadcasts (original tweets), quote tweets (retweet with additional commentary), and replies?\end{tabular} \\ \bottomrule
\end{tabular}%
}
\caption{Sociocultural factors and their corresponding research questions}
\label{tab:sociocultural-questions}
\end{table}





\begin{figure}[h!]
    \centering
    \includegraphics[width=.75\textwidth]{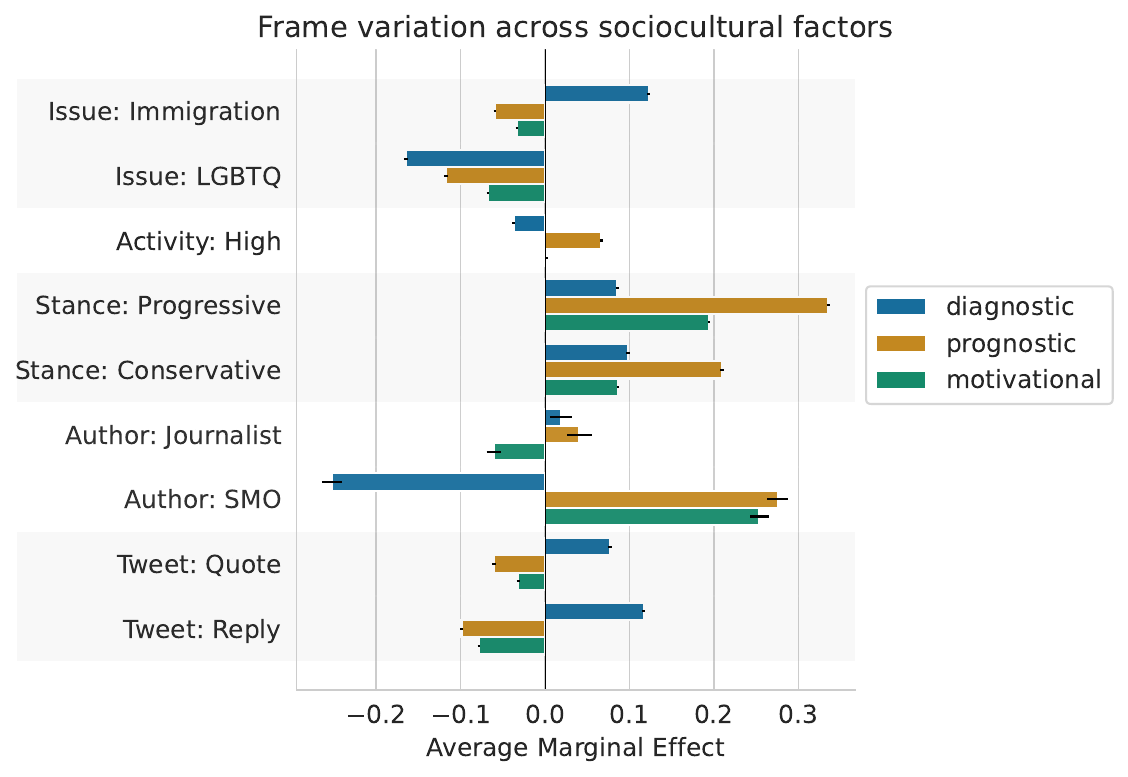}
    \caption{Association between sociocultural factors and framing tasks.}
    \floatfoot{\normalsize \textit{Note.} The x-axis shows average marginal effect estimates from the logistic regression models. Higher values represent stronger associations between sociocultural features and attention to core framing tasks. Error bars represent 95\% confidence intervals.}
    \label{fig:socio_regression_with_stance}
\end{figure}

\subsection{Sociocultural Factor Results}

Logistic regression results are shown in Figure \ref{fig:socio_regression_with_stance} and Table \ref{tab:socio_regression_with_stance}. Results for models excluding stance are qualitatively similar (Appendix Figure \ref{fig:socio_regression_no_stance} and Table \ref{tab:socio_regression_no_stance}). For ease of interpretability, we calculate the average marginal effects of each sociocultural factor for each core framing task. . The average marginal effect represents the average change in probability between the reference level and other values for each sociocultural factor, when all other independent variables are held constant.

\textit{Issue area.} We see substantial cross-issue variation in framing. With all other variables equal, immigration-related tweets are 12.3\% more likely to contain \textit{diagnostic} framing than gun-related tweets but 5.9\% less likely to contain \textit{prognostic} framing and 3.4\% less likely to use \textit{motivational} framing. LGBTQ-related tweets are 16.5\%, 11.8\%, and 6.8\% less likely than gun-related tweets to contain \textit{diagnostic}, \textit{prognostic}, and \textit{motivational} framing, respectively. Table \ref{tab:socio_regression_with_stance} suggests a more nuanced relationship between the issue and prognostic framing in particular. While gun-related tweets are the most likely to contain \textit{prognostic} framing overall, immigration-related tweets are the most likely to discuss \textit{solutions}, and LGBTQ-related tweets are the most likely to express \textit{solidarity} with a movement. While we identify systematic differences across issues, we do not explore the mechanisms underlying these patterns. While it is possible that there are inherent aspects to each issue that contribute to differences in core framing task usage, it is also possible that such variation is due to messages from different issue areas focusing on different types of events and activities.

\textit{Stance.} Relative to tweets with a neutral/unclear stance, tweets with either progressive or conservative stances are much more likely to engage with any core framing task. Conservative tweets are slightly more likely to contain \textit{diagnostic} framing than progressive tweets (9.8\% vs. 8.5\% increase in probability relative to neutral tweets), and particularly the \textit{blame} frame element. Conservative tweets are less likely than progressive tweets to contain \textit{prognostic} framing (21.0\% vs. 33.5\% increase in probability relative to neutral tweets) and \textit{motivational} framing (8.6\% vs. 19.4\% increase in probability). This analysis considers each of the core framing tasks separately. It appears that, overall, stance has stronger associations with core framing tasks than issue. This could be suggestive of alignment within stances (ideologies) in framing strategies across issues. We corroborate this finding with another operationalization, where we measure stance-based alignment by comparing stances in their \textit{distribution} of framing strategies across issues, rather than considering each framing strategy individually (see Appendix for more details).

\textit{Protest activity.} Figure \ref{fig:socio_regression_with_stance} shows that compared to the other sociocultural factors, the protest activity level of the month when a tweet was written has a much weaker relationship with core framing tasks, although still in the expected direction. In months of high protest activity, tweets are 3.7\% less likely to engage in diagnostic framing, 6.7\% more likely to contain prognostic framing, and just 0.3\% more likely to contain motivational framing. Table \ref{tab:socio_regression_with_stance} breaks this down further: high protest activity months are less associated with \textit{problem identification (diagnostic)} and \textit{solution (prognostic)} frame elements, but significantly more associated with \textit{tactics} and \textit{solidarity} prognostic frame elements.

\textit{Author role.} Figure \ref{fig:socio_regression_with_stance} reveals substantial variation in framing strategies across author roles: whether an author is a journalist, SMO, or neither (which we call the ``public''). Journalists are very similar to the public in their use of diagnostic and prognostic framing; in fact, in our regression models that exclude machine-labeled stance (Table \ref{tab:socio_regression_no_stance}), journalists are not significantly different in diagnostic and prognostic framing. Journalists are 6.1\% less likely than the public to use motivational framing. Tweets written by SMOs pattern very differently. Relative to the public, SMO tweets are much less likely to use diagnostic framing (25.2\%), and much more likely to use prognostic and motivational framing (27.5\% and 25.4\%). These effects are surprisingly large, with coefficient estimate magnitudes greater than those for any other sociocultural factor other than stance.


\textit{Tweet type.} Finally, there is variation across tweet types. Compared to broadcast tweets, quote tweets and replies are more likely to contain diagnostic framing by 7.7\% and 11.7\%, respectively. Replies are 9.9\% less likely to contain prognostic framing and 7.9\% less likely to contain motivational framing than broadcasts. Quote tweets show the same pattern but with smaller differences: 6.0\% less likely for  prognostic  and 3.2\% less likely for motivational. The smaller average marginal effects for quote tweets suggest that quote tweets have more in common with broadcasts than replies do. 


\begin{table}[!htbp] \centering 
  \caption{Coefficient estimates from logistic regression models.} 
  \label{tab:socio_regression_with_stance} 
    \resizebox{\linewidth}{!}{
\begin{tabular}{@{\extracolsep{-12pt}}l LLLLLLLL} 
\toprule 
\\[-1.8ex] & \multicolumn{1}{r}{Diagnostic} & \multicolumn{1}{c}{Prognostic} & \multicolumn{1}{r}{Motivational} &
\multicolumn{1}{r}{Identify} & \multicolumn{1}{r}{Blame} & \multicolumn{1}{r}{Solution} & \multicolumn{1}{r}{Tactics} & \multicolumn{1}{r}{Solidarity}\\
\hline \\
 Immigration & 0.629^{***} & -0.261^{***} & -0.280^{***} & 0.786^{***} & 0.498^{***} & 0.636^{***} & -1.272^{***} & -0.572^{***} \\ 
  LGBTQ & -0.703^{***} & -0.518^{***} & -0.627^{***} & -0.369^{***} & -1.144^{***} & -1.343^{***} & -1.791^{***} & 1.412^{***} \\ 
  Conservative & 0.478^{***} & 0.922^{***} & 1.611^{***} & 0.357^{***} & 1.033^{***} & 0.907^{***} & -0.141^{***} & 1.291^{***} \\ 
  Progressive & 0.411^{***} & 1.445^{***} & 2.437^{***} & 0.389^{***} & 0.666^{***} & 1.015^{***} & 1.235^{***} & 3.152^{***} \\ 
  High Activity & -0.188^{***} & 0.291^{***} & 0.024^{***} & -0.178^{***} & -0.007 & -0.133^{***} & 0.346^{***} & 0.676^{***} \\ 
  Journalist & 0.098^{**} & 0.179^{***} & -0.670^{***} & 0.223^{***} & -0.007 & -0.171^{***} & 0.264^{***} & -0.319^{***} \\ 
  SMO & -1.143^{***} & 1.362^{***} & 1.560^{***} & -0.974^{***} & -1.099^{***} & -0.078^{*} & 1.680^{***} & 0.185^{***} \\ 
  Quote & 0.365^{***} & -0.265^{***} & -0.249^{***} & 0.201^{***} & 0.331^{***} & 0.089^{***} & -0.478^{***} & -0.497^{***} \\ 
  Reply & 0.577^{***} & -0.434^{***} & -0.706^{***} & 0.479^{***} & 0.291^{***} & -0.169^{***} & -1.150^{***} & -1.197^{***} \\ 
  \bottomrule
  & \multicolumn{8}{r}{$^{*}$p$<$0.1; $^{**}$p$<$0.05; $^{***}$p$<$0.01} \\ 
\end{tabular} }

\vspace{0.5em}
\RaggedRight{\textit{Note.} Columns represent dependent variables (core framing tasks and frame elements) and rows represent each sociocultural factor. Asterisks show significant coefficients relative to various thresholds after Holm-Bonferroni correction.}
\end{table} 

\begin{figure}[htbp!]
    \centering
    \includegraphics[width=\textwidth]{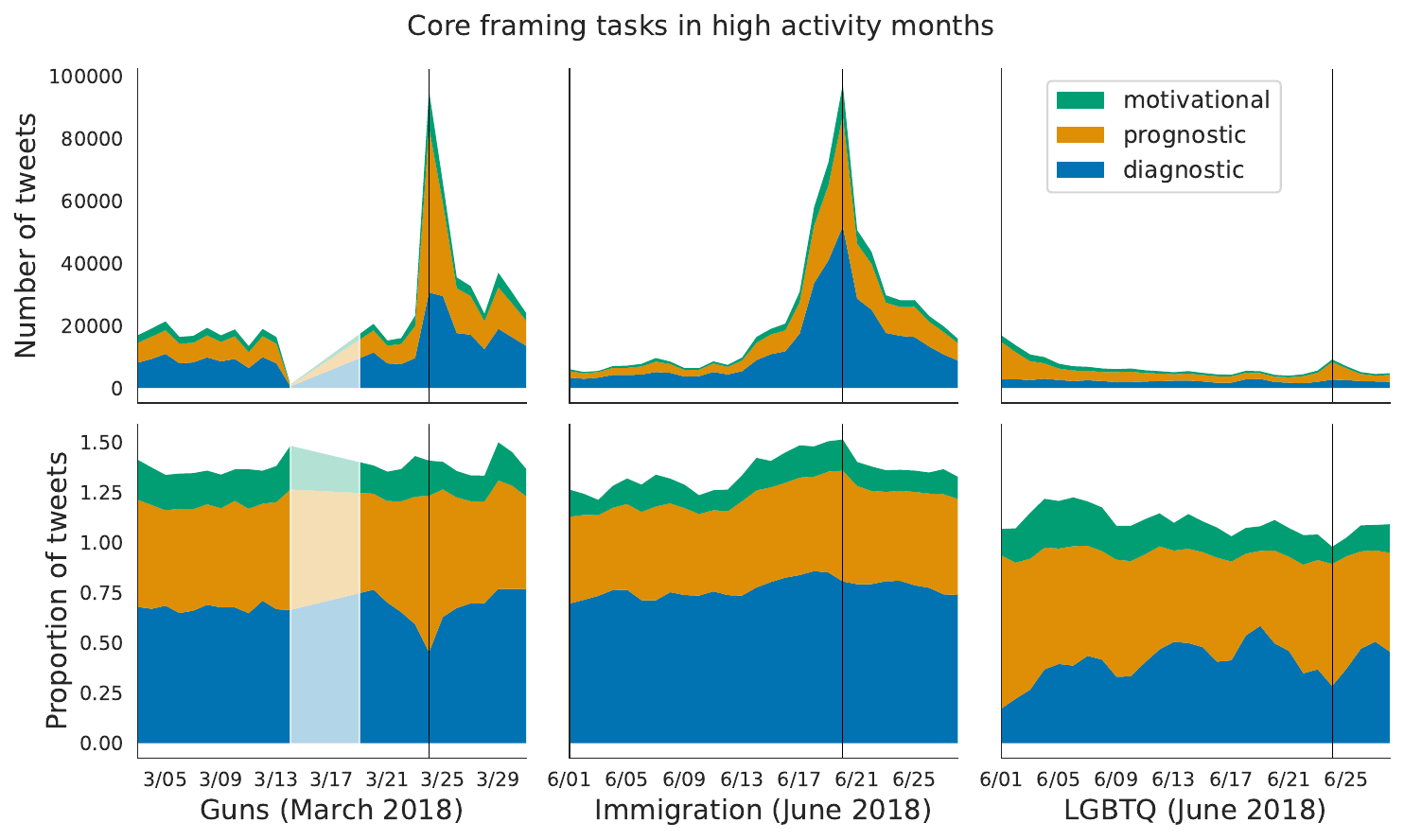}
    \caption{Daily framing task frequency for high protest activity months.}
    \floatfoot{\RaggedRight \normalsize \textit{Note.} Time periods shown are March 2018 for \textit{guns} (left), June 2018 for \textit{immigration} (center) and \textit{LGBTQ rights} (right). Raw counts of core framing tasks are shown in the top row and as a proportion of relevant tweets in the bottom row. The March for Our Lives demonstrations occurred on March 24, shown by the black vertical line in the leftmost plots. The line in the center plots at June 20 is when Trump signed an executive order ending family separation. The line in the rightmost plots occurs at June 24, the date of Pride parades in many U.S. cities including NYC, Chicago, and San Francisco.  Note that the bottom plot does not sum to 1 because core framing tasks are not mutually-exclusive. The translucent gap from 3/14-3/19 is due to missing data from the Decahose stream. Plots also do not include 3/1-3/2 and 6/29-6/30 due to missing data. }
    \label{fig:timeseries_high}
\end{figure}

 \textbf{Fine-grained temporal analysis.}
The sociocultural factor regression analysis reveals surprisingly small framing differences between high and average protest activity months. There are several potential explanations for this result. Perhaps months is too coarse of a timescale to observe framing shifts during protests, which typically occur on a daily scale. Alternatively, maybe the relative attention to \textit{diagnostic}, \textit{prognostic}, and \textit{motivational} framing strategies are truly stable over time regardless of offline protest. To better understand the relationship between time, framing, and protest, we analyze the distribution of core framing tasks for each issue area by day. Figure \ref{fig:timeseries_high} shows temporal patterns for each issue area in their month of high protest activity (see Appendix Figures \ref{fig:timeseries_guns}, \ref{fig:timeseries_immigration}, and \ref{fig:timeseries_lgbtq} for daily temporal plots of high and average activity months side by side for each issue area). 

Here we find evidence for both possible patterns. Within months, and even between low and high activity months, the distribution of \textit{diagnostic}, \textit{prognostic}, and \textit{motivational} framing strategies remain remarkably stable across days. We observe increases in prognostic framing on days with protests, such as March 24, 2018 for \textit{guns} and weekends in June 2018 for \textit{LGBTQ rights} corresponding to major Pride events. Notably, not all spikes in Twitter activity are associated with frame distribution shifts. Unlike the other issues, the spike in activity for immigration does not represent a large collective action event, but rather public outrage and a media storm surrounding family separation at the border. Even as volume exponentially increases during this storm, the balance of framing strategies remains stable. One notable exception to this stability is shown in Figure \ref{fig:timeseries_lgbtq}, which shows distributions of core framing tasks day-by-day for both high and average activity months (June 2018 and April 2019, respectively). During the high activity month, we see considerably higher rates of prognostic framing and lower rates of diagnostic framing for LGBTQ rights in comparison to guns and immigration. However, the average activity month has a much higher proportion of diagnostic framing, with the distribution more strongly resembling the other issues. The sustained increased prognostic framing throughout the entire high activity month, but only for LGBTQ rights, suggests that the distinct nature of Pride may shape our findings about both issue area and protest activity levels.

Finally, we ask: which stakeholder group is participating the most in these framing changes during protest days: journalists, SMOs, or the public? While we cannot fully answer this question due to much sparser data for journalists and SMOs, Figure \ref{fig:timeseries_stakeholders} in the Appendix shows daily distributions of framing strategies for these groups. Interestingly, while SMOs generally engage in much higher levels of prognostic framing than the general public, their use of prognostic framing does not increase during protest days. Rather, SMOs increase their motivational framing in the days preceding major protests, such as March for Our Lives. We speculate that around these events, the division of labor shifts between movement actors: the general public (or perhaps rank-and-file activists) take on the work of prognostic framing, while SMOs focus on motivational framing and other movement activities beyond the symbolic dimension.



\section{Discussion}

Framing is an active, dynamic, and contested process \citep{benford_framing_2000} that is not an inherent property of social movements. It develops through the assemblage of conversations between stakeholders such as activists, bystanders, opponents, organizations, and the media.  In the digital age characterized by personalization and connective action \citep{bennett_digital_2011,bennett_logic_2012}, where political action is more dispersed and decentralized \citep{kavada_social_2016}, it is important to consider the content, context, speakers, and audiences of individual messages in the study of framing \citep{earl_new_2017}. Beyond challenging earlier understandings of collective action, social media offers a unique window into understanding the development of social movement frames not just at the movement-level, but at the level of millions of individual messages \citep{kavada_social_2016}.   

To better understand social movement mobilization on social media, we develop a computational approach to study collective action framing based on \citet{snow_ideology_1988}'s typology of \textit{diagnostic}, \textit{prognostic}, and \textit{motivational} core framing tasks. By creating a codebook, a manually-labeled dataset of 6,000 tweets, and sophisticated machine learning models, we infer framing strategies for nearly two million tweets across three issue areas and multiple time periods. This approach enables us to conduct the empirical comparative work lacking in extant social movement scholarship, particularly in framing \citep{tarrow1996social,snow_emergence_2014}; our large-scale data facilitates analyses of frame variation not only across issues, but also stance, protest activity levels, author roles, and tweet interaction types.   

Our analysis begins with an investigation into how tweets make use of lower-level linguistic resources to accomplish higher-level core framing tasks. While many linguistic features associated with each core framing task are issue-specific, we identify several consistent themes across all three issue areas. For example, we identify moral language to be prominent in both diagnostic and prognostic framing, and 3rd person pronouns to be highly associated with diagnostic framing, suggestive of boundary framing processes \citep{hunt1994identity}. From the vantage point of computational methods, we reconstruct the centrality of collective ingroup identity in social movement discourse \citep{polletta2001collective}, drawing connections between social movement studies and social psychology for future work to further build upon. Beyond our analysis of linguistic features, our fine-grained temporal analysis of core framing tasks also reveals remarkable stability in the relative proportions or diagnostic, prognostic, and motivational framing strategies on a day-by-day basis. We hope future work further delves into this tension between the stability and dynamism of framing; what gives rise to such consistent patterns, and how can we reconcile these perspectives?

We then continue with an analysis of how framing varies across five sociocultural factors: issue area, stance, protest activity, author role, and tweet interaction type. Each part of this analysis speaks both to ongoing debates in social movement scholarship and sheds light on potential directions for future work. For example, our day-by-day temporal analyses show shifts in relative proportions of each core framing task during protest days (i.e. \textit{March for Our Lives} and \textit{Pride parades}) but not during other notable but non-protest days with spikes in Twitter activity (i.e. in the (social) media storm surrounding family separation at the US-Mexico border). We do not know if this pattern generalizes across different types of offline events, and we do not attempt to determine if the type of offline event has any causal effect on online framing or vice versa. Nevertheless, these patterns highlight the potential--and a starting point--for further work to delve deeper into the connection between social media framing and offline events. 

Our data provides us with a unique opportunity to understand the discursive construction of meaning in online social networks. We showcase these processes through our analysis of author roles and tweet types, highlighting that different kinds of participants and different kinds of interactions offer different kinds of meanings. While journalists and the general public engage with each core framing task to a similar degree, SMOs behave extremely differently, with far more \textit{prognostic} and \textit{motivational} framing than the other groups. Prior work has questioned and debated the relevance of SMOs in the digital age \citep{earl_future_2015,bozarth_beyond_2021}. While we cannot provide a definitive answer about the importance of SMOs based on this descriptive analysis alone, these substantial differences suggest that SMOs at least play a unique role in the online social movement ecosystem. Focusing on tweet types, we show that ``broadcast'' tweets are much more likely to cue prognostic and motivational strategies, while both replies and quote tweets are more likely to cue diagnostic strategies. This finding emphasizes that social media meaning-making occurs not through one-sided messaging, but through conversations, with each kind of interaction offering a unique contribution to the broader discourse. This analysis of tweet types begins to explore the relationship between framing and the affordances of social media platforms. Future research can unpack this relationship more: how do platform affordances affect collective action framing? Do differences in affordances across platforms shape how social movement mobilization strategies?

As this is solely a descriptive study, we do not attempt to make any arguments regarding framing effects. However, such description lays foundations for future research to address a broad range of causal questions. For example, what are the effects of exposure to diagnostic, prognostic, and motivational framing strategies? Does framing impact individual audience members' participation in a social movement, or their perceptions of the movement's scope, efficacy, or necessity? Beyond perceptions, does framing actually affect movement participation and success? Especially given our finding that SMOs engage in more motivational framing immediately preceding protests, we may ask if framing drives protest activity. Future work could also measure if certain kinds of framing strategies tend to receive more engagement on social media \citep{mendelsohn2021modeling}. Higher engagement could be suggestive of stronger effects on audiences and lead to increased amplification by social media recommender systems, thus having major implications for the role of platforms in shaping social movement outcomes.

In addition to these directions for theoretically-motivated future work, we identify many opportunities for methodological innovation. We use a specific set of keywords to collect data from the Twitter 10\% sample from just three issue areas, within which we consider only two months of activity from between 2018-2019. We do not know if our computational models or the patterns uncovered in our analyses are generalizable to other issues, time periods, or platforms. For example, do progressives always use more \textit{prognostic} framing than conservatives? Or do we observe this pattern because progressive movements were more active in this time period, and may have been more likely to use Twitter for organizing? Do gun-related movements generally discuss \textit{tactics} more than immigration-related movements, or is that merely a reflection of the fact that there were unprecedented demonstrations for gun control in 2018? 

Particularly in light of Twitter and other platforms restricting data access for researchers and limiting opportunities for curating new social media datasets, this question is more urgent than ever: how can we ensure that the tools we build for large-scale social movement analysis work effectively for different movements, time periods, and platforms? We anticipate that transfer learning modeling approaches would be necessary here. While our dataset is limited, it contains three distinct issue areas each with two different months, and can thus provide a good testbed for future transfer learning model evaluation. 

Future work may also consider integrating generative large language models (LLMs) such as ChatGPT into computational frame analysis. Recent work has considered LLMs as a tool to replace or augment human-provided annotations for many social science tasks \citep{ziems2023can}, including news credibility rating \citep{yang_large_2023} and political affiliation prediction \citep{tornberg_chatgpt-4_2023}. In a set of data labeling tasks that closely resemble ours, \citet{gilardi_chatgpt_2023} finds that ChatGPT outperforms crowdworkers in labeling tweets about content moderation for relevance, stance, topics, media policy frames, and distinguishing tweets that frame content moderation as a problem vs. as a solution. However, the validity of LLM annotations has not yet been established in our domain of collective action framing on social media. Moreover, LLMs still perform considerably worse than trained experts for text annotation \citep{gilardi_chatgpt_2023} and present additional concerns of reliability and consistency \citep{reiss_testing_2023}. We suggest that LLMs should not be used in a fully unsupervised manner, but rather in collaboration with humans \citep{reiss_testing_2023,ziems2023can}.

Theoretically, our descriptive study can serve as a foundation for future work to measure how framing strategies and frame \textit{processes} (e.g. frame bridging, transformation, extension, and resonance) unfold through a complex, dynamic network of interactions on social media \citep{snow_frame_1986, jost_how_2018}. This work opens avenues for empirical research to explore how framing affects political, cultural, biographical, and other dimensions of social movement success. Methodologically, we demonstrate the utility of computational methods for social movement content analysis and identify specific opportunities for computational techniques in subsequent research. As \citet{snow_emergence_2014} argue, ``empirical investigations of framing hold the potential to influence activists’ practice toward greater efficacy in mobilizing recruits and gaining media attention'', suggesting that our work can hold direct implications for activist practices and strategies.


\section{Acknowledgments}
We thank audiences at Text as Data 2023 (TADA) and Politics and Computational Social Science 2023 (PaCSS) for their helpful feedback on an earlier version of this work. We also thank the anonymous reviewers for their insightful comments and suggestions. This work was supported by the National Science Foundation (Grant IIS-1815875). C.B. thanks the Center for Advanced Study in Behavioral Sciences. J.M. gratefully acknowledges support from the Google PhD Fellowship.

\bibliography{ref}
\newpage

\appendix
\section{Online Appendix}
\label{sec:appendix}
\setcounter{table}{0}
\setcounter{figure}{0}

\renewcommand{\thetable}{A\arabic{table}}
\renewcommand{\thefigure}{A\arabic{figure}}

\subsection{Model Performance}
\label{sec:appendix_model}

\begin{table}[htbp!]
\centering
\resizebox{.75\textwidth}{!}{%
\begin{tabular}{@{}llll@{}}
\toprule
\textbf{Issue} & \textbf{Liberal} & \textbf{Neutral/Unclear} & \textbf{Conservative} \\ \midrule
Guns           & 0.731 (0.045)    & 0.539 (0.060)            & 0.631 (0.074)         \\
Immigration    & 0.766 (0.015)    & 0.661 (0.041)            & 0.757 (0.013)         \\
LGBTQ          & 0.855 (0.015)    & 0.588 (0.040)            & 0.183 (0.129)         \\ \bottomrule
\end{tabular}%
}
\caption{Development F1-scores of stance classification}

\vspace{0.5em}
\RaggedRight{\textit{Note.} Separated by issue, averaged over five cross-validation folds. Standard deviation values are shown in parentheses.}
\label{tab:stance_dev}
\end{table}

\subsection{Linguistic Analysis}
\label{sec:appendix_linguistic}

\begin{figure}[htbp!]
    \centering
    \includegraphics[width=.5\textwidth]{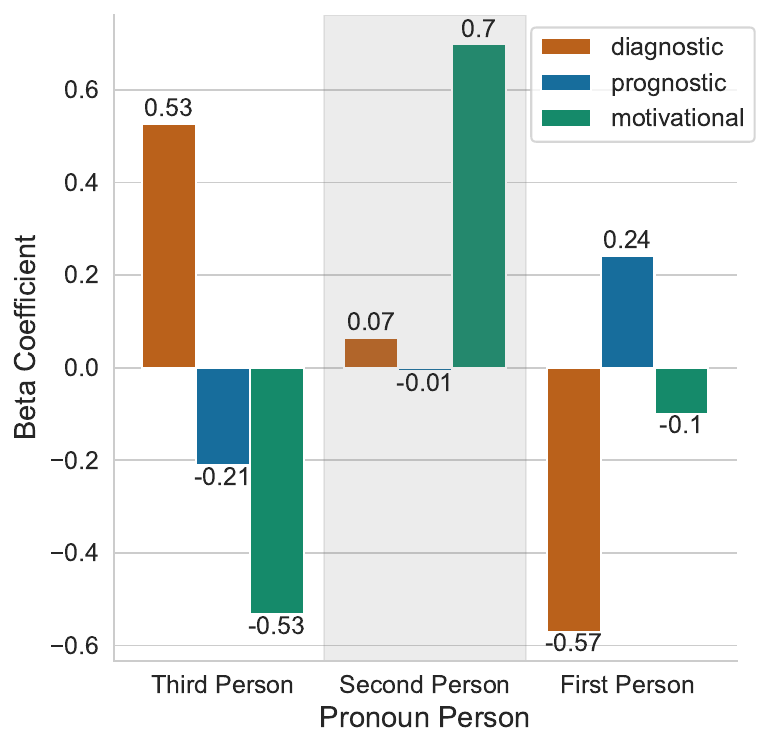}
    \caption{Associations between core framing tasks and pronoun person.}
    \floatfoot{\RaggedRight \normalsize \textit{Note.}  The y-axis represents $\beta$ coefficient estimates, where higher values represent stronger positive associations. Units of analysis are pronouns, dependent variables are pronoun person marking, and independent variables include issue area and whether the tweet in which the pronoun appears contains diagnostic, prognostic, or motivational framing strategies. Diagnostic framing is most associated with 3rd person pronouns, prognostic framing is most  associated with 1st person pronouns, and motivational framing is most associated with 2nd person pronouns.}
    \label{fig:pronouns}
\end{figure}

\subsection{Sociocultural Factors}
\label{sec:appendix_sociocultural}

\begin{table}[!htbp] \centering 
    \caption{Coefficients from logistic regression models excluding stance.} 
  \label{tab:socio_regression_no_stance} 
    \resizebox{\linewidth}{!}{
\begin{tabular}{@{\extracolsep{-12pt}}l LLLLLLLL} 
\toprule 
\\[-1.8ex] & \multicolumn{1}{r}{Diagnostic} & \multicolumn{1}{c}{Prognostic} & \multicolumn{1}{r}{Motivational} &
\multicolumn{1}{r}{Identify} & \multicolumn{1}{r}{Blame} & \multicolumn{1}{r}{Solution} & \multicolumn{1}{r}{Tactics} & \multicolumn{1}{r}{Solidarity}\\
\hline \\
 Immigration & 0.613^{***} & -0.309^{***} & -0.362^{***} & 0.767^{***} & 0.481^{***} & 0.587^{***} & -1.289^{***} & -0.660^{***} \\ 
  LGBTQ & -0.689^{***} & -0.280^{***} & -0.313^{***} & -0.338^{***} & -1.178^{***} & -1.246^{***} & -1.402^{***} & 1.734^{***} \\ 
  High Activity & -0.168^{***} & 0.353^{***} & 0.131^{***} & -0.157^{***} & 0.008^{*} & -0.087^{***} & 0.425^{***} & 0.788^{***} \\ 
  Journalist & 0.003 & -0.042 & -0.814^{***} & 0.146^{***} & -0.210^{***} & -0.344^{***} & 0.224^{***} & -0.441^{***} \\ 
  SMO & -1.058^{***} & 1.716^{***} & 1.967^{***} & -0.879^{***} & -1.015^{***} & 0.148^{***} & 2.072^{***} & 0.732^{***} \\ 
  Quote & 0.435^{***} & -0.044^{***} & -0.022^{**} & 0.264^{***} & 0.468^{***} & 0.251^{***} & -0.332^{***} & -0.303^{***} \\ 
  Reply & 0.638^{***} & -0.327^{***} & -0.638^{***} & 0.520^{***} & 0.453^{***} & -0.052^{***} & -1.187^{***} & -1.211^{***} \\ 
  \bottomrule
  & \multicolumn{8}{r}{$^{*}$p$<$0.1; $^{**}$p$<$0.05; $^{***}$p$<$0.01} \\ 
\end{tabular} }

\vspace{0.5em}
\RaggedRight{\textit{Note.} Columns represent dependent variables (core framing tasks and frame elements) and rows represent each sociocultural factor. Asterisks show significant coefficients relative to various thresholds after Holm-Bonferroni correction.}
\end{table}

\begin{figure}[htbp!]
    \centering
    \includegraphics[width=.75\textwidth]{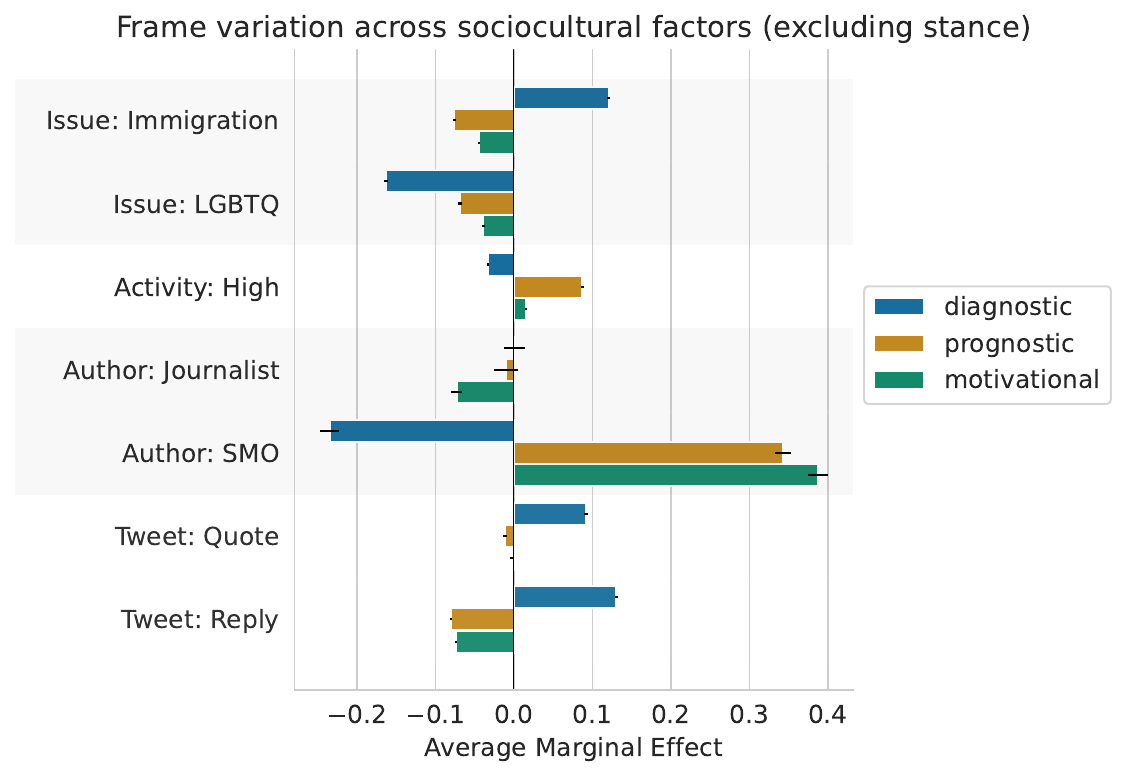}
    \caption{Associations between sociocultural factors and core framing tasks excluding stance.}
    \floatfoot{\RaggedRight \normalsize \textit{Note.} The x-axis represents average marginal effect estimates for each factor from the logistic regression models. Higher values represent stronger associations between sociocultural features and attention to core framing tasks. Error bars represent 95\% confidence intervals.}
    \label{fig:socio_regression_no_stance}
\end{figure}

Given our findings from issue area and stance, we further ask: are ideologically-similar messages from different issue areas more aligned in framing strategies than ideologically-opposed messages from the same issue area? We answer this by calculating the pairwise relative entropy of the distributions of core framing tasks and frame elements between four groups of tweets: progressive gun-related, conservative gun-related, progressive immigration-related, and conservative immigration-related (we exclude LGBTQ-related tweets from this analysis due to the low performance in detecting anti-LGBTQ tweets). Lower relative entropy between two groups means that their distributions of framing strategies are more similar, and thus have greater frame alignment. We calculate entropy using 1,000 bootstrapped samples of 10,000 tweets each. We find that alignment is highest (i.e. lowest entropy) in framing across issues within the same ideology (progressive = 0.049, conservative = 0.052) and lowest between opposing ideologies within the same issue area (guns = 0.09, immigration = 0.097)

\begin{figure}[htbp!]
    \centering
    \includegraphics[width=.7\textwidth]{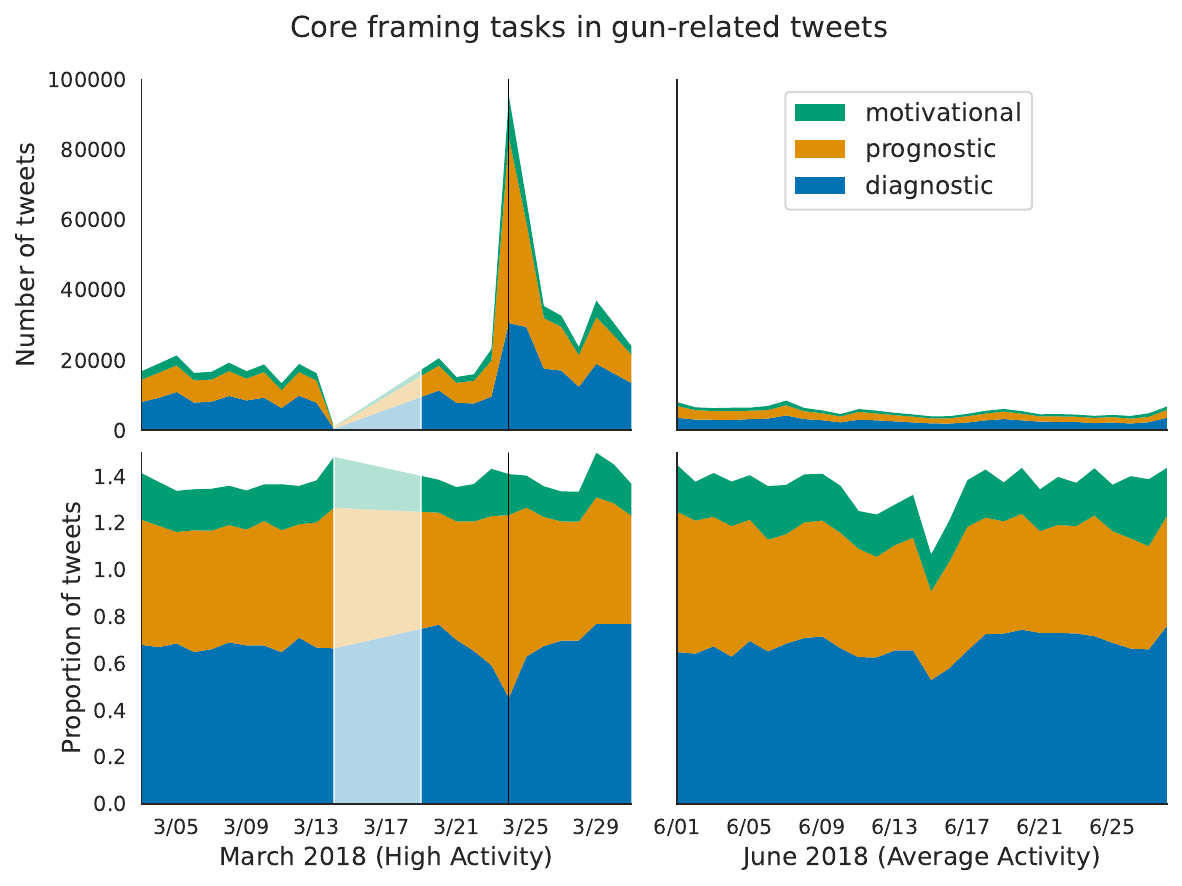}
    \caption{Core framing tasks per day for guns.}
    \label{fig:timeseries_guns}
\end{figure}

\begin{figure}[htbp!]
    \centering
    \includegraphics[width=.7\textwidth]{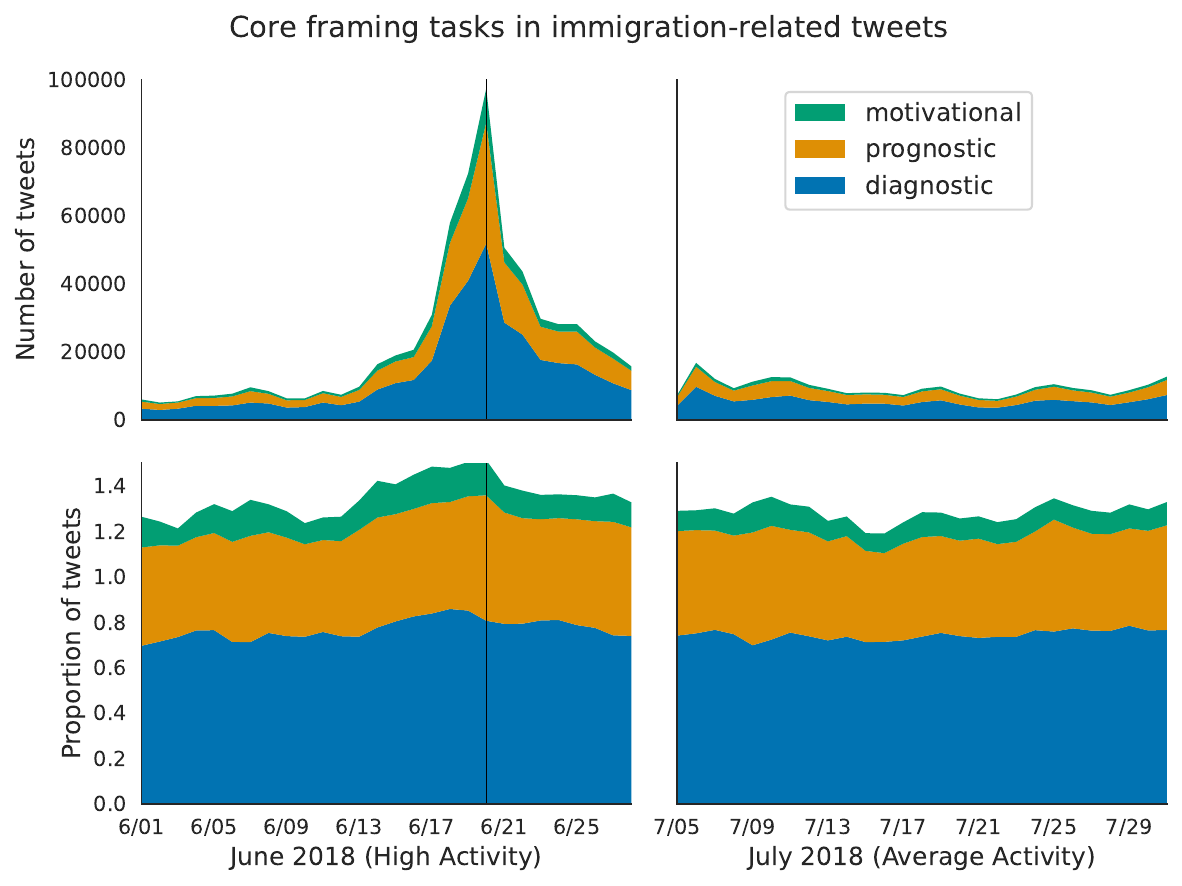}
    \caption{Core framing tasks per day for immigration}
    \label{fig:timeseries_immigration}
\end{figure}

\begin{figure}[htbp!]
    \centering
    \includegraphics[width=.7\textwidth]{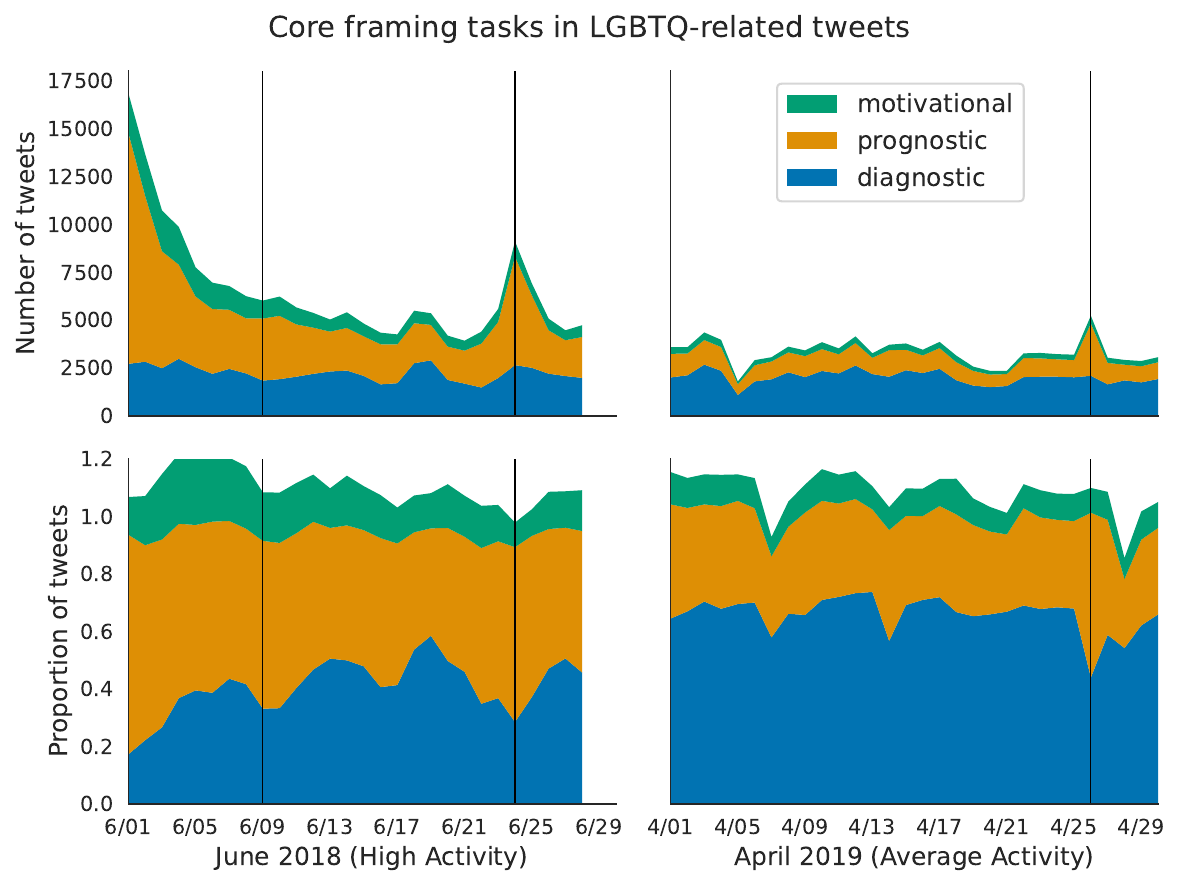}
    \caption{Core framing tasks per day for LGBTQ rights.}
    \floatfoot{\RaggedRight \normalsize \textit{Note.} Vertical lines in June 2018 represent days with many Pride parades. The vertical line at April 26, 2019 represents Lesbian Visibility Day.}
    \label{fig:timeseries_lgbtq}

\end{figure}

\begin{figure}[htbp!]
    \centering
    \includegraphics[width=\textwidth]{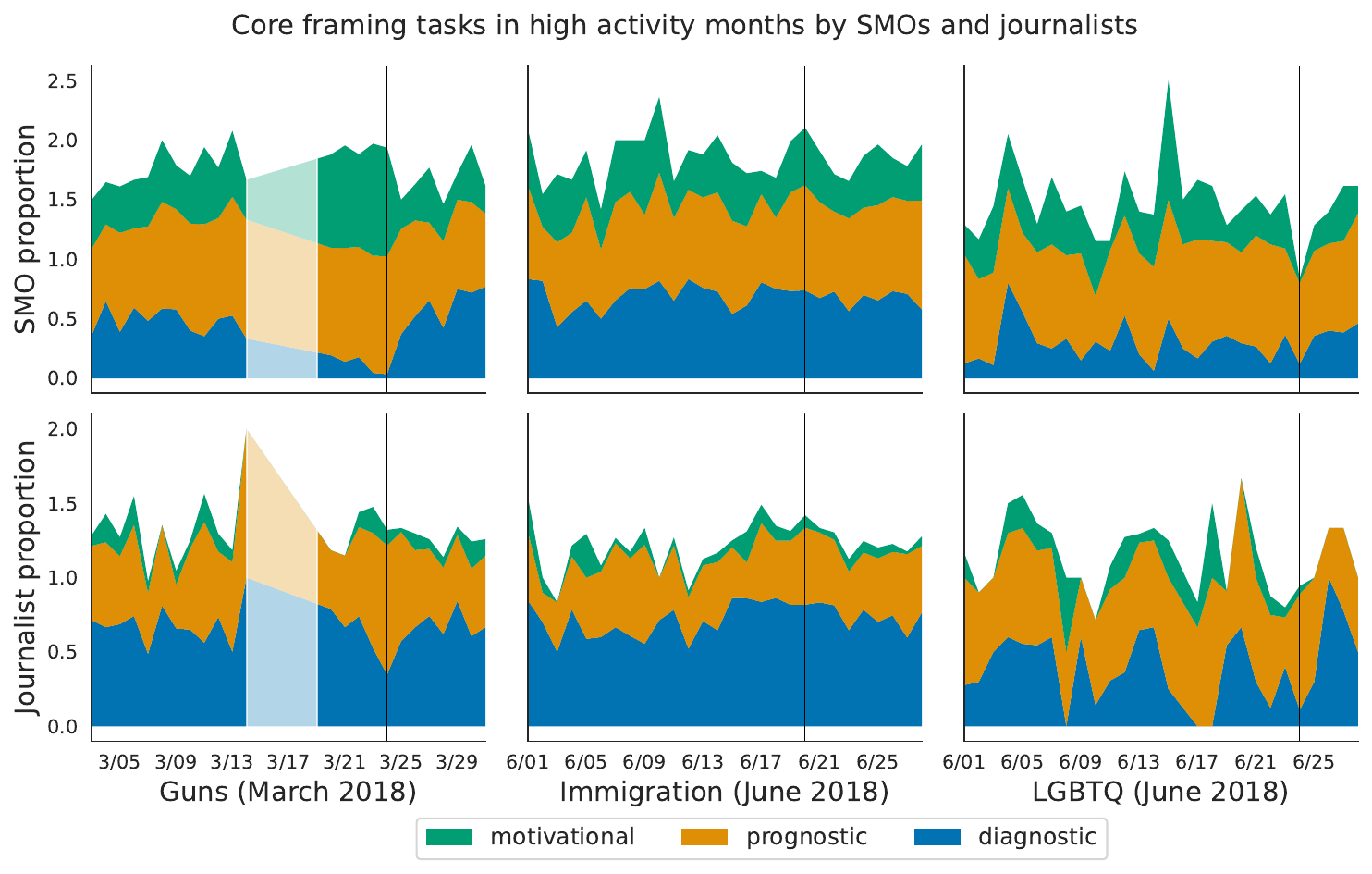}
    \caption{Core framing task frequency per day in high activity months for journalists and social movement organizations (SMOs).}
    \label{fig:timeseries_stakeholders}
\end{figure}

\begin{table}[htbp!]
\centering
\resizebox{\textwidth}{!}{%
\begin{tabular}{|l|lll|lll|lll|} \hline
 & 
  \multicolumn{3}{c|}{Diagnostic} &
  \multicolumn{3}{c|}{Prognostic} &
  \multicolumn{3}{c|}{Motivational} \\ \hline
\multirow{16}{*}{Words} &
  Guns &
  Immigration &
  LGBTQ &
  Guns &
  Immigration &
  LGBTQ &
  Guns &
  Immigration &
  LGBTQ \\ \midrule
 &
  shooting &
  children &
  homophobic &
  marchforourlives &
  stop &
  pride &
  stop &
  please &
  please \\
 &
  school &
  separation &
  homophobia &
  march &
  \begin{tabular}[c]{@{}l@{}}keepfamilies-\\ together\end{tabular} &
  happy &
  backfiretrump &
  stop &
  twibbon \\
 &
  they &
  illegally &
  anti &
  today &
  should &
  [rainbow emoji] &
  bloodshed &
  join &
  help \\
 &
  fatal &
  parents &
  that &
  stop &
  buildthewall &
  support &
  suffering &
  sign &
  add \\
 &
  after &
  their &
  people &
  we &
  we &
  help &
  potus &
  your &
  support \\
 &
  suffering &
  trump &
  they &
  neveragain &
  need &
  twibbon &
  fatal &
  petition &
  now \\
 &
  backfiretrump &
  stop &
  racist &
  gun &
  to &
  month &
  m14 &
  you &
  pastel \\
 &
  that &
  policy &
  is &
  backfiretrump &
  abolishice &
  add &
  today &
  help &
  bisexual \\
 &
  bloodshed &
  separated &
  not &
  control &
  wall &
  please &
  entered &
  \begin{tabular}[c]{@{}l@{}}familiesbe-\\ longtogether\end{tabular} &
  lgbt \\
 &
  stop &
  kids &
  he &
  bloodshed &
  must &
  now &
  jra &
  stand &
  pride \\
 &
  is &
  is &
  hate &
  suffering &
  deport &
  love &
  @sootch00 &
  need &
  stop \\
 &
  he &
  democrats &
  against &
  potus &
  end &
  pastel &
  longguns &
  call &
  a \\
 &
  killed &
  child &
  do &
  vote &
  \begin{tabular}[c]{@{}l@{}}keepfamiles-\\ together\end{tabular} &
  winagun &
  @moveon &
  join \\
 &
  their &
  they &
  transphobia &
  change &
  thank &
  bisexual &
  guncontest &
  support &
  your \\
 &
  potus &
  \begin{tabular}[c]{@{}l@{}}trumpconcen-\\ trationcamps\end{tabular} &
  n't &
  our &
  worldrefugeeday &
  [green heart] &
  @classicfirearm &
  portman &
  our \\ \hline
\multirow{15}{*}{Verbs} &
  kill &
  separate &
  be &
  stop &
  stop &
  help &
  stop &
  stop &
  help \\
 &
  suffer &
  stop &
  do &
  suffer &
  should &
  support &
  suffer &
  join &
  add \\
 &
  stop &
  kill &
  hate &
  march &
  need &
  add &
  longgun &
  need &
  support \\
 &
  shoot &
  cross &
  say &
  vote &
  must &
  celebrate &
  winagun &
  help &
  stop \\
 &
  attack &
  break &
  call &
  need &
  thank &
  love &
  enter &
  sign &
  join \\
 &
  die &
  blame &
  stop &
  thank &
  join &
  thank &
  unsubscribe &
  stand &
  let \\
 &
  blame &
  lie &
  misgender &
  ban &
  help &
  join &
  sign &
  donate &
  check \\
 &
  lose &
  rip &
  refuse &
  join &
  end &
  need &
  join &
  demand &
  click \\
 &
  murder &
  refuse &
  kill &
  should &
  abolishice &
  stop &
  need &
  support &
  need \\
 &
  be &
  enter &
  attack &
  will &
  sign &
  hope &
  vote &
  must &
  epub \\
 &
  bully &
  commit &
  homophobic &
  sign &
  worldrefugeeday &
  fight &
  let &
  elect &
  learn \\
 &
  do &
  lose &
  should &
  support &
  build &
  should &
  register &
  let &
  donate \\
 &
  try &
  destroy &
  use &
  protest &
  demand &
  share &
  stay &
  please &
  spread \\
 &
  fail &
  care &
  deny &
  end &
  resist &
  will &
  dreamgun &
  wake &
  visit \\
 &
  ignore &
  force &
  defend &
  fight &
  stand &
  donate &
  theme &
  read &
  sign \\ \hline
\multirow{15}{*}{Adjectives} &
  fatal &
  bad &
  homophobic &
  fatal &
  safe &
  happy &
  fatal &
  vulnerable &
  pastel \\
 &
  unarmed &
  racist &
  anti &
  proud &
  buildthewall &
  proud &
  guncont &
  dear &
  more \\
 &
  black &
  inhumane &
  racist &
  enough &
  well &
  amazing &
  sweet &
  historic &
  free \\
 &
  bad &
  cruel &
  transphobic &
  young &
  open &
  beautiful &
  involved &
  more &
  excellent \\
 &
  mass &
  illegal &
  bad &
  common &
  simple &
  pastel &
  official &
  @cbp &
  new \\
 &
  dead &
  criminal &
  lgbt &
  safe &
  \begin{tabular}[c]{@{}l@{}}familiesbe-\\ longtogether\end{tabular} &
  great &
  vast &
  safe &
  online \\
 &
  white &
  evil &
  sexist &
  amazing &
  great &
  safe &
  strong &
  @credomobile &
  sissy \\
 &
  innocent &
  sick &
  white &
  sensible &
  proud &
  wonderful &
  enough &
  current &
  hashtag \\
 &
  mental &
  wrong &
  wrong &
  powerful &
  full &
  inclusive &
  young &
  inhumane &
  safe \\
 &
  sick &
  dangerous &
  religious &
  strict &
  adequate &
  important &
  @libertymutual &
  immigrant &
  available \\
 &
  violent &
  disgusting &
  disgusting &
  strong &
  healthy &
  more &
  dear &
  urgent &
  next \\
 &
  illegal &
  human &
  misogynistic &
  beautiful &
  strong &
  excited &
  common &
  analytic &
  criminal \\
 &
  disgusting &
  innocent &
  gay &
  more &
  humane &
  awesome &
  preventable &
  reunite &
  important \\
 &
  dangerous &
  homeless &
  hateful &
  great &
  more &
  \begin{tabular}[c]{@{}l@{}}happypride-\\ month\end{tabular} &
  ashamed &
  @rpfranceue &
  open \\
 &
  russian &
  dead &
  violent &
  gunreformnow &
  executive &
  fellow &
  near &
  @speakerryan &
  civil \\ \hline
\multirow{15}{*}{\begin{tabular}[c]{@{}l@{}}Subject-\\ Verb\end{tabular}} &
  potus\_stop &
  they\_want &
  i\_hate &
  potus\_stop &
  we\_need &
  i\_love &
  potus\_stop &
  we\_need &
  we\_need \\
 &
  police\_shoot &
  they\_care &
  you\_say &
  we\_need &
  future\_await &
  we\_need &
  i\_enter &
  i\_sign &
  mobi\_epub \\
 &
  texas\_suffer &
  what\_happen &
  they\_’re &
  reply\_stop &
  we\_want &
  we\_celebrate &
  \begin{tabular}[c]{@{}l@{}}dreamgun-\\ \_longgun\end{tabular} &
  people\_elect &
  you\_need \\
 &
  they\_want &
  democrats\_want &
  i\_understand &
  i\_sign &
  letter\_’ &
  i\_hope &
  reply\_stop &
  we\_elect &
  you\_want \\
 &
  people\_die &
  they\_break &
  that\_’ &
  i\_march &
  i\_sign &
  we\_love &
  texas\_suffer &
  you\_reverse &
  people\_need \\
 &
  ohio\_suffer &
  who\_break &
  i\_agree &
  we\_want &
  they\_need &
  mobi\_epub &
  we\_need &
  million\_want &
  you\_like \\
 &
  who\_lose &
  who\_cross &
  he\_’ &
  i\_support &
  we\_elect &
  i\_vote &
  ohio\_suffer &
  \begin{tabular}[c]{@{}l@{}}refugee-\\ \_deserve\end{tabular} &
  you\_help \\
 &
  police\_kill &
  you\_care &
  \begin{tabular}[c]{@{}l@{}}brunei-\\ \_implement\end{tabular} &
  voice\_hear &
  people\_elect &
  we\_stand &
  we\_register &
  us\_do &
  you\_do \\
 &
  illinois\_suffer &
  you\_break &
  who\_think &
  texas\_suffer &
  i\_stand &
  i\_support &
  illinois\_suffer &
  you\_do &
  ’s\_keep \\
 &
  who\_survive &
  immigrant\_kill &
  \begin{tabular}[c]{@{}l@{}}you-\\ \_homophobic\end{tabular} &
  i\_stand &
  problem\_solve &
  we\_have &
  voice\_hear &
  you\_need &
  you\_stop \\
 &
  that\_kill &
  they\_try &
  they\_do &
  we\_march &
  \begin{tabular}[c]{@{}l@{}}globalcompact-\\ migration\_reflect\end{tabular} &
  everyone\_have &
  \begin{tabular}[c]{@{}l@{}}marchforourl-\\ ive\_unfold\end{tabular} &
  you\_help &
  you\_join \\
 &
  they\_care &
  this\_happen &
  he\_do &
  we\_register &
  refugee\_deserve &
  people\_need &
  i\_sign &
  \begin{tabular}[c]{@{}l@{}}parliament-\\ \_indicate\end{tabular} &
  you\_have \\
 &
  nra\_own &
  illegal\_kill &
  they\_hate &
  ohio\_suffer &
  we\_stop &
  swift\_donate &
  california\_suffer &
  pls\_act &
  we\_do \\
 &
  that\_protect &
  dem\_want &
  they\_think &
  illinois\_suffer &
  we\_have &
  we\_support &
  louisiana\_suffer &
  you\_join &
  i\_urge \\
 &
  california\_suffer &
  it\_’ &
  you\_hate &
  we\_have &
  you\_reverse &
  racists\_homophobic &
  that\_protect &
  you\_have &
  we\_celebrate \\ \hline
\multirow{15}{*}{\begin{tabular}[c]{@{}l@{}}Verb-\\ Object\end{tabular}} &
  stop\_bloodshed &
  break\_law &
  have\_right &
  stop\_bloodshed &
  thank\_you &
  add\_twibbon &
  stop\_bloodshed &
  sign\_petition &
  add\_twibbon \\
 &
  kill\_people &
  cross\_border &
  use\_slur &
  thank\_you &
  build\_wall &
  support\_pride &
  winagun\_m14 &
  elect\_you &
  support\_pride \\
 &
  do\_nothing &
  separate\_child &
  say\_thing &
  sign\_petition &
  sign\_petition &
  support\_pastel &
  enter\_contest &
  add\_name &
  support\_pastel \\
 &
  take\_money &
  separate\_family &
  make\_comment &
  end\_violence &
  stop\_separation &
  thank\_you &
  sign\_petition &
  join\_portman &
  join\_we \\
 &
  shoot\_man &
  take\_child &
  hate\_people &
  find\_march &
  deport\_they &
  love\_you &
  find\_march &
  \begin{tabular}[c]{@{}l@{}}stop-\\ \_separation\end{tabular} &
  try\_something \\
 &
  kill\_kid &
  do\_nothing &
  say\_what &
  make\_change &
  await\_child &
  celebrate\_pride &
  win\_308abatan &
  take\_action &
  pass\_equalityact \\
 &
  survive\_shooting &
  use\_child &
  have\_problem &
  ban\_weapon &
  thank\_@hp &
  join\_we &
  end\_violence &
  do\_something &
  show\_love \\
 &
  push\_agenda &
  enter\_country &
  refuse\_service &
  save\_life &
  use\_tech &
  see\_you &
  join\_today &
  join\_we &
  like\_this \\
 &
  kill\_man &
  commit\_crime &
  call\_i &
  join\_today &
  connect\_they &
  try\_something &
  winagun\_ar15 &
  join\_i &
  support\_border \\
 &
  blame\_nra &
  separate\_kid &
  make\_fun &
  stop\_violence &
  end\_separation &
  show\_support &
  theme\_rifle &
  tell\_congress &
  tell\_congress \\
 &
  take\_right &
  put\_child &
  call\_they &
  raise\_age &
  keep\_family &
  raise\_awareness &
  win\_flag &
  pass\_amnesty &
  need\_right \\
 &
  stop\_shooting &
  stop\_separation &
  misgender\_i &
  vote\_they &
  elect\_you &
  support\_you &
  save\_child &
  support\_bill &
  spread\_word \\
 &
  do\_anything &
  rip\_child &
  \begin{tabular}[c]{@{}l@{}}end-\\ \_homophobia\end{tabular} &
  pass\_legislation &
  add\_name &
  celebrate\_pridemonth &
  follow\_we &
  \begin{tabular}[c]{@{}l@{}}deserve-\\ \_shelter\end{tabular} &
  \begin{tabular}[c]{@{}l@{}}educate-\\ \_yourself\end{tabular} \\
 &
  save\_child &
  separate\_they &
  call\_he &
  \begin{tabular}[c]{@{}l@{}}support-\\ \_marchforourlive\end{tabular} &
  secure\_border &
  celebrate\_diversity &
  pass\_legislation &
  slam\_door &
  \begin{tabular}[c]{@{}l@{}}make-\\ \_homophobia\end{tabular} \\
 &
  kill\_child &
  add\_name &
  make\_joke &
  \begin{tabular}[c]{@{}l@{}}watch-\\ \_marchforourlive\end{tabular} &
  take\_action &
  spread\_love &
  take\_min &
  reverse\_it &
  support\_circle \\ \hline
\end{tabular}%
}
\caption{Top 15 linguistic features for each core framing task by issue. }
\label{tab:logodds}
\end{table}

\end{document}